\documentclass[10pt,twocolumn]{article} 

\usepackage{IEEE}
\usepackage{amsmath,amssymb}
\usepackage{newtxtext,newtxmath}

\usepackage{microtype}
\usepackage[english]{babel}
\usepackage{titlesec}

\usepackage{cite}

\usepackage{algorithmic}
\usepackage{graphicx}
\usepackage{textcomp}
\usepackage[table]{xcolor}  
\usepackage{booktabs, makecell, booktabs,multirow} 
\usepackage{url} 
\usepackage[colorlinks]{hyperref}
\usepackage{caption}
\usepackage{subcaption}
\usepackage{pifont}
\usepackage{tikz}
\usepackage{IEEEtrantools}
\usepackage[symbol, hang,flushmargin]{footmisc}

\newcommand{\etal}{\textit{et al.}\ }
\newcommand{\ie}{\textit{i.e.,}\ }
\newcommand{\eg}{\textit{e.g.,}\ }

\newcolumntype{L}[1]{>{\raggedright\let\newline\\\arraybackslash\hspace{0pt}}m{#1}}
\newcolumntype{C}[1]{>{\centering\let\newline\\\arraybackslash\hspace{0pt}}m{#1}}
\newcolumntype{R}[1]{>{\raggedleft\let\newline\\\arraybackslash\hspace{0pt}}m{#1}}

\usepackage{stackengine}

\definecolor{emerald}{rgb}{0.31, 0.78, 0.47}
\definecolor{OliveGreen}{rgb}{0,0.7,0}
\definecolor{caribbeangreen}{rgb}{0.0, 0.8, 0.6}
\definecolor{ao(english)}{rgb}{0.0, 0.5, 0.0}
\definecolor{kellygreen}{rgb}{0.3, 0.73, 0.09}

\let\oldding\ding
\renewcommand{\ding}[2][1]{\scalebox{#1}{\oldding{#2}}}

\usepackage{pdfpages}

\begin{document}


\title{An Overview of Recent Work in Media
Forensics: Methods and Threats}


\author{
Kratika Bhagtani, Amit Kumar Singh Yadav, Emily R. Bartusiak,\\
Ziyue Xiang, Ruiting Shao, Sriram Baireddy, Edward J. Delp \vspace{0.5em}\\ 
\normalsize Video and Image Processing Laboratory (VIPER)\\
\normalsize School of Electrical and Computer Engineering\\
\normalsize Purdue University \\
\normalsize West Lafayette, Indiana, USA
}

\maketitle
\thispagestyle{plain}

\begin{abstract}

In this paper, we review recent work in media forensics for digital images, video, audio (specifically speech), and documents.
For each data modality, we discuss  synthesis and manipulation techniques that can be used to create and modify digital media.
We then review technological advancements for detecting and quantifying such manipulations.
Finally, we consider open issues and suggest directions for future research. 

\end{abstract}

\renewcommand*{\thefootnote}{\arabic{footnote}}

\section{Introduction}\label{part-1-intro}


The acquisition and circulation of digital media (\ie images, video, and audio) has become popular with the proliferation of digital capture devices (\eg smartphones), free tools for editing and manipulating digital media content (\eg GIMP~\cite{gimp}), and social networks. 
With developments in deep learning~\cite{goodfellow2016deep, hastie2001elements}, manipulating digital media and generating synthetic content has become very easy~\cite{perov2020deepfacelab, shen2018natural}.
The manipulations can look extremely realistic and be challenging to detect~\cite{farid2009seeing}.
The intent behind such manipulations is important to consider.
For example, media could be manipulated to create misinformation or to commit financial fraud~\cite{nightingale2022aisynthesized,wakefield2022deepfake}. 
Attempts to use manipulated multimedia for influencing social discourse, elections, and the way people interact in a civilized society~\cite{vaccari2021deepfakes} have driven research efforts in multimedia forensics~\cite{lyu2021fighting,battiato2016multimedia}.

Multimedia forensics is the area of research that includes signal processing, computer vision, machine learning, statistics, psychology, sociology, and political science to study the development of manipulated and synthetic media and methods that can be used to detect and mitigate its use.
The goals of media forensics are to answer the following questions: \textit{Is the media element manipulated (detection)?}; \textit{Where is it modified (localization)?}; \textit{What tools and/or who modified it (attribution)?}; and \textit{Why did they modify it (characterization)?}
In this paper we focus on detection, localization, and attribution.

Multimedia signals (\eg image, video, audio) have many characteristics that can be analyzed for detection, attribution, and localization.
Some methods analyze pixels to determine if and how images and videos are manipulated.
Audio methods can analyze waveform amplitudes and frequencies.
Other methods utilize information about the ``construction'' of a digital asset.
All digital assets are created by a system consisting of acquisition and in-device processing.
For example, a typical digital camera system consists of acquisition through a system of lenses and in-camera processing such as Color Filter Array (CFA) interpolation, white balance, and gamma correction~\cite{piva2013overview, nguyen2013counterforensics}. Finally, the output is compressed.
These processing methods leave traces---or fingerprints---in the information of a media asset that reveals the acquisition device and processing methods~\cite{farid2009image, delp2005multimedia}. 
We can attribute the media to its acquisition system by analyzing these fingerprints~\cite{chen2008determining,khanna2007scanner, shao2020forensic,khanna2007sensor,cozzolino2019extracting}. 
Additionally, we can use the fingerprints to verify whether the media is pristine, manipulated, or synthesized~\cite{chen2008determining}.
When an attacker manipulates the media, the fingerprints can be affected, enabling detection and localization of the manipulations.
Deep learning models used to generate media, such as Generative Adversarial Networks (GANs)~\cite{gui2021review} and diffusion models~\cite{dhariwal2021diffusion}, also have fingerprints.
Thus, media forensics methods can also strive to recognize fingerprints of the generation method as part of their analysis~\cite{marra2019do, gragnaniello2021are}. 

There are several review papers in media forensics, including image forensics~\cite{fridrich2009digital,piva2013overview,yang2020survey}, video forensics~\cite{bestagini2012overview, javed2021comprehensive, shelke2020comprehensive, cozzolino2022multimedia, verdoliva2020media}, audio forensics~\cite{zakariah2017digital}, theoretical analysis~\cite{moulin2005datahiding,chu2016information,kraetzer2015considerations}, author attribution~\cite{rocha2016authorship}, and document forensics~\cite{saini2016forensic,chiang2009printer,doermann2010evolution}.
In this paper we review recent work in media forensics for images, video, audio, and documents and emphasize work in machine learning approaches.
Note this is a longer version of a paper presented at the 2022 IEEE International Conference on Multimedia Information Processing and Retrieval entitled ``An Overview of Recent Work in Multimedia Forensics''~\cite{bhagtani2022overview}.



\section{Image Forensics}\label{part-2-image}



\begin{figure}[t]
    \centering
    \includegraphics[width=0.47\textwidth]{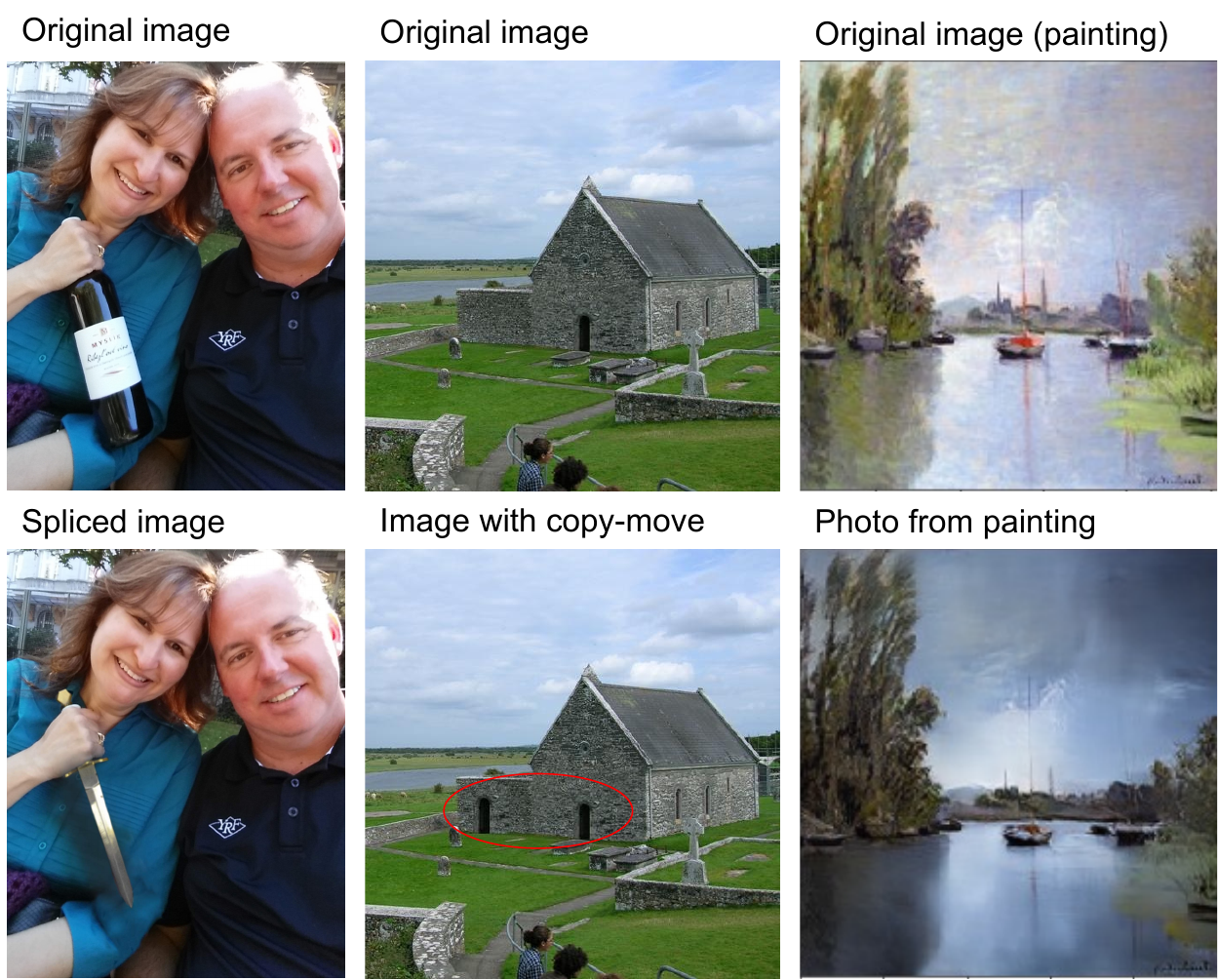}
    \caption{Examples of image manipulations. Images (taken from the MFC Dataset\protect\footnotemark) on the left show splicing. The top row shows the original image, which has been spliced to form the bottom image by replacing the wine bottle with dagger. Images in the middle (taken from the MICC Dataset\protect\footnotemark) show copy-move manipulation. The door from the top original image has been copied and pasted in another location in the bottom manipulated image. Images on the right (synthetically generated by CycleGAN) show style transfer from a painting (top) to a realistic-looking photo (bottom)~\cite{zhu2017unpaired}.}
    \label{image_manipulations}
\end{figure}

In this paper, we discuss images that are considered to be \textit{unaltered}, \textit{manipulated}, or \textit{synthesized}.
Unaltered images have been neither manipulated nor synthesized; they are authentic images as captured by a camera system.
Typically, synthesized images refer to entire images that have been generated from scratch.
Manipulated images are images in which portions of the images have been altered.
Note that the manipulated portion of the image could could be synthesized. 
Some common examples of image manipulation techniques include splicing~\cite{nataraj2021holistic} (replacing a section of an image with a section from another image), inpainting~\cite{yang2017highresolution} (deleting a section of an image and synthesizing pixels to replace the content), copy-move~\cite{yu2020manipulation} (duplicating a section of an image and moving it to another position within the same image), and photo-montage~\cite{tan2018where} (composite image from a combination of two or more images).
Figure~\ref{image_manipulations} shows examples of splicing and copy-move manipulations. 


The level of realism that synthetic images have attained poses challenges in distinguishing them from unaltered images.
This is complemented by counter forensic approaches that use strategies to deceive forensic methods. 
For example, G\"{u}era \etal added adversarial noise---specific types of noise that are indiscernible to the human eye and known to fool Convolutional Neural Networks (CNNs)---into images.
The authors then showed that CNN-based camera model attribution methods were negatively impacted~\cite{guera2017a}.
Bonettini \etal proposed a method to imperceptibly alter an image by removing camera-specific noise to severely hinder sensor noise-based camera model detectors~\cite{bonettini2018fooling}.
Cozzolino \etal injected traces of real cameras into synthetic images to deceive detectors into identifying them as real~\cite{cozzolino2021spoc}.
Huang \etal proposed a method to evade detectors via notch filtering in the spatial domain~\cite{huang2021dodging}. 
These approaches further motivate multimedia forensics research.

In this section, we first introduce some recent methods for image manipulation and synthetic image generation, and then discuss  developments in  detection methods.

\footnotetext[1]{\url{https://www.nist.gov/itl/iad/mig/media-forensics-challenge-2018}}
\footnotetext{\url{http://lci.micc.unifi.it/labd/2015/01/copy-move-forgery-detection-and-localization/}}

\subsection{Image Manipulation and Synthesis}

Common image manipulations such as splicing, copy-move attack, seam-carving, and inpainting can be performed using constraint-based methods~\cite{barnes2009patchmatch} and deep learning~\cite{goodfellow2016deep, hastie2001elements}.
Several methods have been proposed for automatic image manipulation. CNNs were described in~\cite{tan2018where} for image compositing~\cite{schetinger2016digital}.
Yang \etal and Shetty \etal demonstrated automatic image inpainting utilizing a multi-scale neural network~\cite{yang2017highresolution} and Generative Adversarial Network (GAN)~\cite{goodfellow2014generative}, respectively. 
The methods proposed for image inpainting sometimes produce blurry regions or artifacts.
This lack of high-frequency information can make manipulation detection easier.
More recent methods aim to generate sharper edges of inpainted regions~\cite{nazeri2019edgeconnect}.

There has also been significant progress in research related to synthetic image generation~\cite{goodfellow2014generative,dhariwal2021diffusion}. 
GANs~\cite{goodfellow2014generative} have enabled the synthesis of high-quality image content that is visually almost indistinguishable from unaltered images. 
A GAN typically consists of two competing  networks: a generator that attempts to learn a data distribution and a discriminator that attempts to distinguish synthetic data created by the generator from the original data it is learning to model~\cite{goodfellow2014generative,gui2021review}.
GANs have been used to translate styles of an image collection (\eg paintings) into an unrelated image collection~\cite{zhu2017unpaired}.
Figure~\ref{image_manipulations} shows an example of a style transfer from a painting to a photograph.
Karras \etal described a training method for progressive growth of GANs to stabilize training and progressively improve the quality of generated images~\cite{karras2018progressive}. 


\begin{figure}[htpb]
    \centering
    \includegraphics[width=0.48\textwidth]{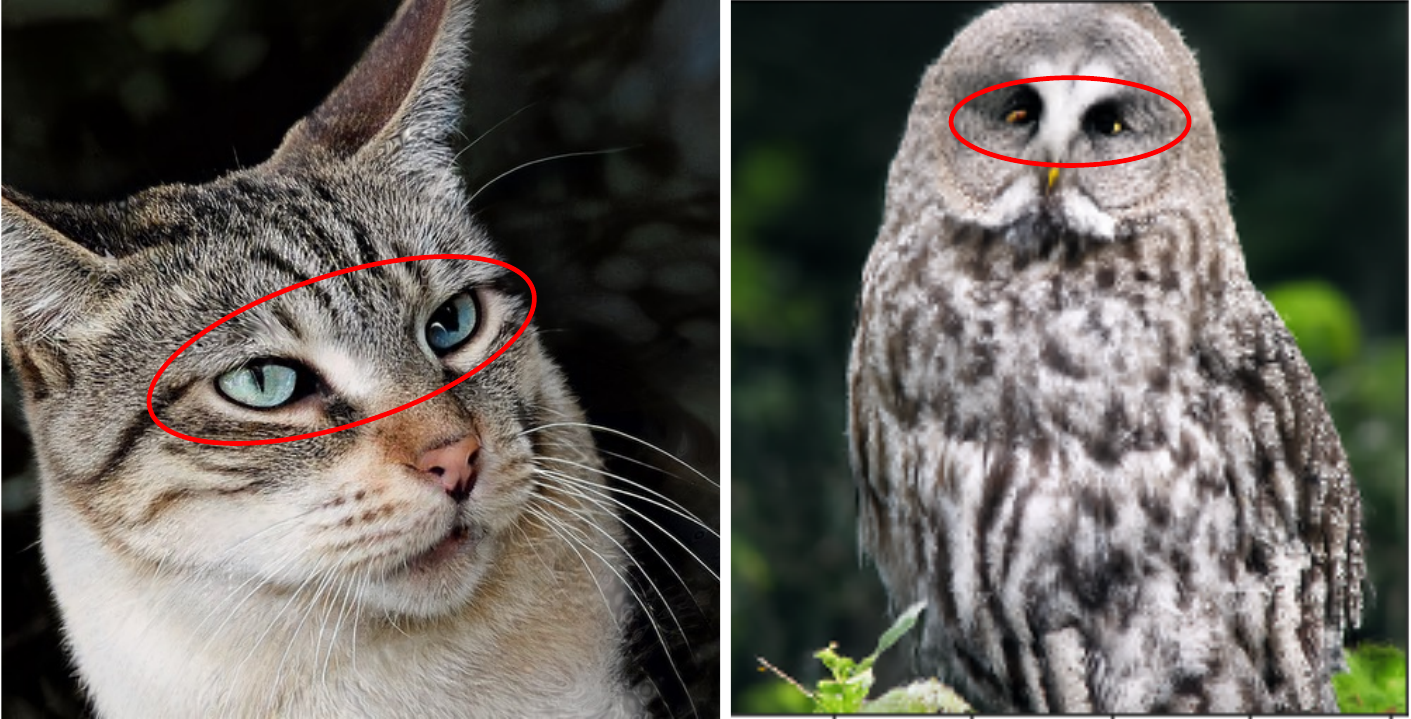}
    \caption{Synthetic images from a GAN (left) and a Diffusion model (right). 
    Both images have mismatched pupil shape and eye color.
    The image on the left is generated by StyleGAN3~\cite{karras2021stylegan}; the image on the right is generated by~\cite{dhariwal2021diffusion}.
    }
    \label{gan_inconsistent}
\end{figure}

A style-based generator (\eg StyleGAN~\cite{karras2019stylebased}) can control image synthesis by adjusting the style of a learned image at each convolution layer.
In recent years, attempts have been made to improve the quality of GAN-generated images and address characteristic artifacts through improvements in architecture and training strategies~\cite{karras2020analyzing,nagano2019deep}.
Brock \etal described a real-world class-conditional image synthesis method trained on complex datasets that yielded high-fidelity images~\cite{brock2019large}. 
Karras \etal proposed a high-quality synthetic image generation method trained with limited data (\ie with only a few thousand training images)~\cite{karras2020training}. 
In~\cite{karras2021aliasfree}, an alias-free GAN known as StyleGAN3 was proposed, and the generating process is equivariant to translation and rotation.
Figure~\ref{gan_inconsistent} shows a synthetic image generated using this alias-free GAN.

Besides GANs, there are other image synthesis methods with the potential to generate high-quality synthetic images. 
Likelihood-based diffusion models~\cite{dhariwal2021diffusion} and score-based generative models~\cite{vahdat2021scorebased,song2021scorebased} can also create synthetic images with high-frequency information, including sharp edges and fine details. 
Figure~\ref{gan_inconsistent} shows a synthetic image generated using a likelihood-based diffusion model~\cite{dhariwal2021diffusion}.
Chan \etal utilized 3D GANs for the synthesis of multi-view images of faces and animals~\cite{chan2021efficient}.
Transformers, originally developed for natural language processing, can also generate high resolution images~\cite{esser2021taming}. 
DALL-E~\cite{ramesh2021zero, ramesh2022hierarchichal} is a text-to-image method that uses a transformer based on GPT-3~\cite{gpt3}.
It receives a text description as an input and produces images that fit that description.
The rapid improvement in synthetic image generation methods demands equally robust detection methods.

\subsection{Image Manipulation Detection}



Many methods have been proposed to detect image splicing, which is a common type of manipulation.
Wu \etal used a Deep Matching and Validation Network~\cite{wu2017deep}. 
The network estimates a probability that a potential donor image has been used to splice a given query image and generates splicing masks for both images.
One drawback of this approach is that along with the spliced query image, it also requires the donor image, which may not always be available.
Nataraj \etal used a CNN-based method that operates on pixel co-occurrence matrices for image manipulation detection~\cite{nataraj2021holistic}, which only requires the spliced query image.
First,  the co-occurrence matrices of the image under analysis are computed.
The co-occurrence matrices describe textures within the image and their locations using a histogram of pixel pair values.
Then, a ResNet50~\cite{he2016deep} network analyzes the co-occurrence matrices and determines if the image is manipulated. 
This method is effective for a variety of manipulation techniques, but it does not provide a localization map indicating where the manipulations occur within the image.

Liu \etal proposed a method for detection and localization that utilizes both a matching strategy and an adversarial strategy~\cite{liu2019adversarial}. 
In this approach, two images are analyzed to detect whether regions in each of the input images are the same. 
The method involves three networks: a deep-matching network, a detection network, and a discriminative network. 
The deep-matching network generates manipulation masks for both input images, highlighting matching regions.
This part of the network serves as the generator of the adversarial approach.
The detection network predicts whether the two images with their manipulation masks are correlated or not.
The role of the discriminator network is to drive the generator to produce manipulation masks that are consistent with their ground truths.
Although this method provides extra localization information about manipulation compared to the work proposed by Nataraj \etal~\cite{nataraj2021holistic}, one drawback is that it  requires two input images. 
More specifically, it requires the image that was used to splice the region into the host image, which might not always be available.

Other adversarial approaches do not employ a matching technique and thus do not require the image from which the spliced region is taken.
For example, a conditional Generative Adversarial Network (cGAN) can be used to detect and localize forgeries in images~\cite{bartusiak2019splicing, bartusiak2019adversarial, yarlagadda2019shadow, yarlagadda2018satellite}.
In these approaches, a generator-discriminator pair is still utilized, but the generator only operates on a single input image.
Thus, the generator is forced to detect manipulations based on inherent artifacts in the input image, without relying on a matching network or images that may contain the spliced pixels.

Some media forensics methods detect physical inconsistencies in images to determine if the images are unaltered. 
For example, Yarlagadda \etal and Kumar \etal investigated shadows~\cite{yarlagadda2019shadow, kumar2019image}.
Zhu \etal decomposed face images into their 3D geometry and lighting parameters~\cite{zhu2021face}.
Other methods use media information, such as JPEG compression artifacts, for manipulation detection and localization~\cite{niu2021image, kwon2021catnet, yu2020manipulation}. 
Bonettini \etal ~\cite{bonettini2019image} proposed a manipulation detection method that focuses on the absence of camera sensor noise in images. 
They showed that CNNs trained with this feature could generalize better in terms of unseen devices.
G\"{u}era \etal analyzed which regions in images were most suitable for attribution detection~\cite{guera2018reliability}.

Other techniques use an ensemble of methods to authenticate an image~\cite{charitidis2021operationwise, barni2021copy}. 
Charitidis \etal used five detection and localization methods: two are based on detecting JPEG compression using Discrete Cosine Transform (DCT), another two are based on JPEG compression in spatial domain, and one is based on noise fingerprint approach. The decisions from these five methods were fused using deep learning for image manipulation detection and localization~\cite{charitidis2021operationwise}.
Barni \etal proposed a multi-branch CNN to localize tampered areas in a copy-move attack by identifying the source and target regions~\cite{barni2021copy}.

\begin{figure}[htpb]
    \centering
    \includegraphics[width=0.48\textwidth]{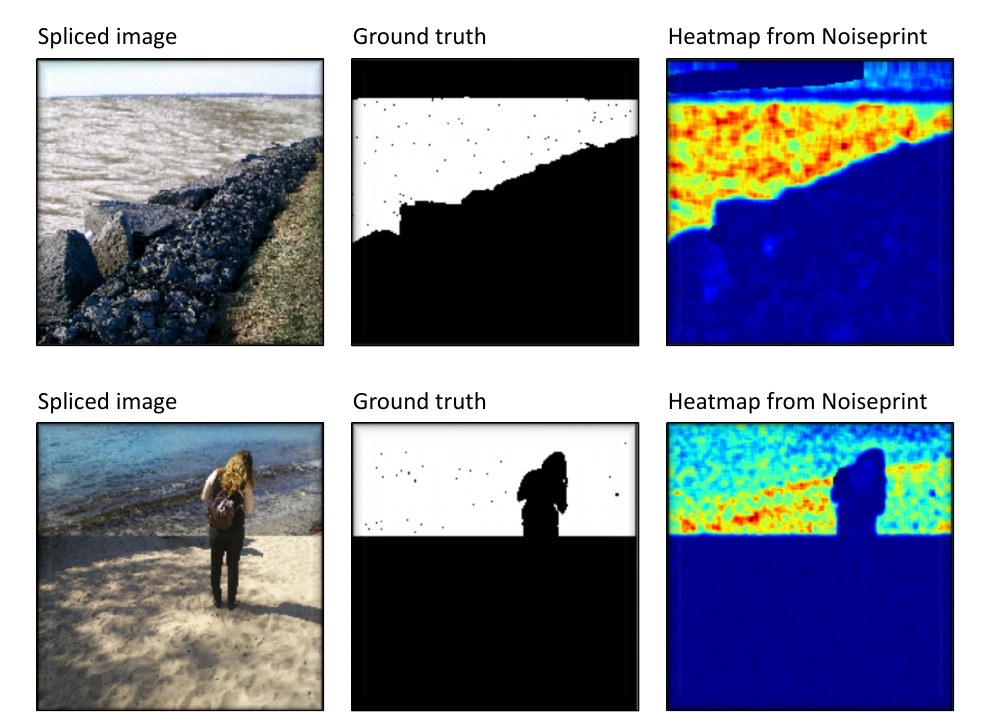}
    \caption{Splicing detection using Noiseprint~\cite{cozzolino2019noiseprint}. Both rows show spliced images (left) taken from MFC Dataset\protect\footnotemark, their ground truths (middle), and the corresponding heatmaps generated from Noiseprint (right). In the ground truth images, the white region highlights the manipulation.}
    \label{image_forgery}
\end{figure}
\footnotetext{\url{https://www.nist.gov/itl/iad/mig/media-forensics-challenge-2018}}

In realistic scenarios, manipulated images contain a combination of manipulations. 
There are methods that work to detect~\cite{nataraj2021holistic} and localize~\cite{wu2019mantranet,chen2021image} manipulations in such situations. 
Noiseprint, a camera model fingerprint using a CNN, was proposed in~\cite{cozzolino2019noiseprint} for image manipulation localization. 
The method produced heatmaps with brighter regions suggesting a possible manipulation.
Figure~\ref{image_forgery} shows two examples of splicing and the corresponding heatmaps generated using Noiseprint for splicing detection.

The methods described above, which work well for detecting manipulations in ``consumer camera images'', sometimes require more effort before they can be applied to satellite images.
Satellites usually have different camera sensors and acquisition methods, which can create different images than consumer cameras.
For example, Cannas \etal developed methods for satellite attribution based on panchromatic imagery~\cite{cannas2021open}.
Horv\'{a}th \etal proposed splicing detection methods in satellite images using a vision transformer~\cite{horvath2021manipulation}, a deep belief network~\cite{horvath2020manipulation}, and an autoencoder combined with a one-class classifier~\cite{horvath2019manipulation}.
Montserrat \etal used generative auto-regressive models to detect manipulations when their nature is unknown, which is the case in realistic scenarios~\cite{montserrat2020generative}.

\subsection{Synthetic Image Detection}

Methods for synthetic image detection often involve GAN-specific solutions which assume prior knowledge of the generative process.
Giudice \etal proposed a solution by detecting anomalous frequencies while analyzing DCT coefficients of the image~\cite{giudice2021fighting}.
This work also showed that GANs leave architecture-specific fingerprints/signatures, which are not directly perceivable but are present in the spatial frequency domain of the synthetically generated image.

Recently, one of the focus points in image forensics has been detecting whether the image is synthetic without prior knowledge of its source or its manipulation history.
Methods capable of accomplishing this level of detection are helpful in real-life scenarios.
Several methods focused on identifying the GAN from which the synthetic image is generated. 
Wang \etal~\cite{wang2020cnngenerated}, Girish \etal~\cite{girish2021towards} and Cozzolino \etal~\cite{cozzolino2021towards} demonstrated that training a classifier for a specific GAN generator could also be generalized for other GANs by detecting common synthetic image inconsistencies.
Guarnera \etal used the detection of convolutional traces in synthetic images as a method of revealing them, which showed good performance for multiple GAN models~\cite{guarnera2020fighting}.
Detection of synthetic images generated using GANs has been reviewed in detail~\cite{gragnaniello2021are}.

Inconsistencies in GAN-generated images which occur due to the absence of any physiological constraints in the generative process can be used for detection. 
For example, Hu \etal, and Guo \etal utilized corneal specular highlights~\cite{hu2021exposing} and irregularity in pupil shapes~\cite{guo2021eyes}, respectively, to reveal synthetic GAN faces. 
They exploited the idea that when eyes in real faces look straight into the camera, they will see the same scene. 
Also, pupil shapes should be symmetrical for humans, which is not always consistent in synthetic images. Figure~\ref{gan_inconsistent} shows examples of synthetic images with physical inconsistencies.
This line of thought can also be extended to detect semantic inconsistencies to expose synthetic images (\eg asymmetrical eyes and mismatched earrings).
\begin{figure*}[htpb]
    \centering
    \includegraphics[width=0.8\textwidth]{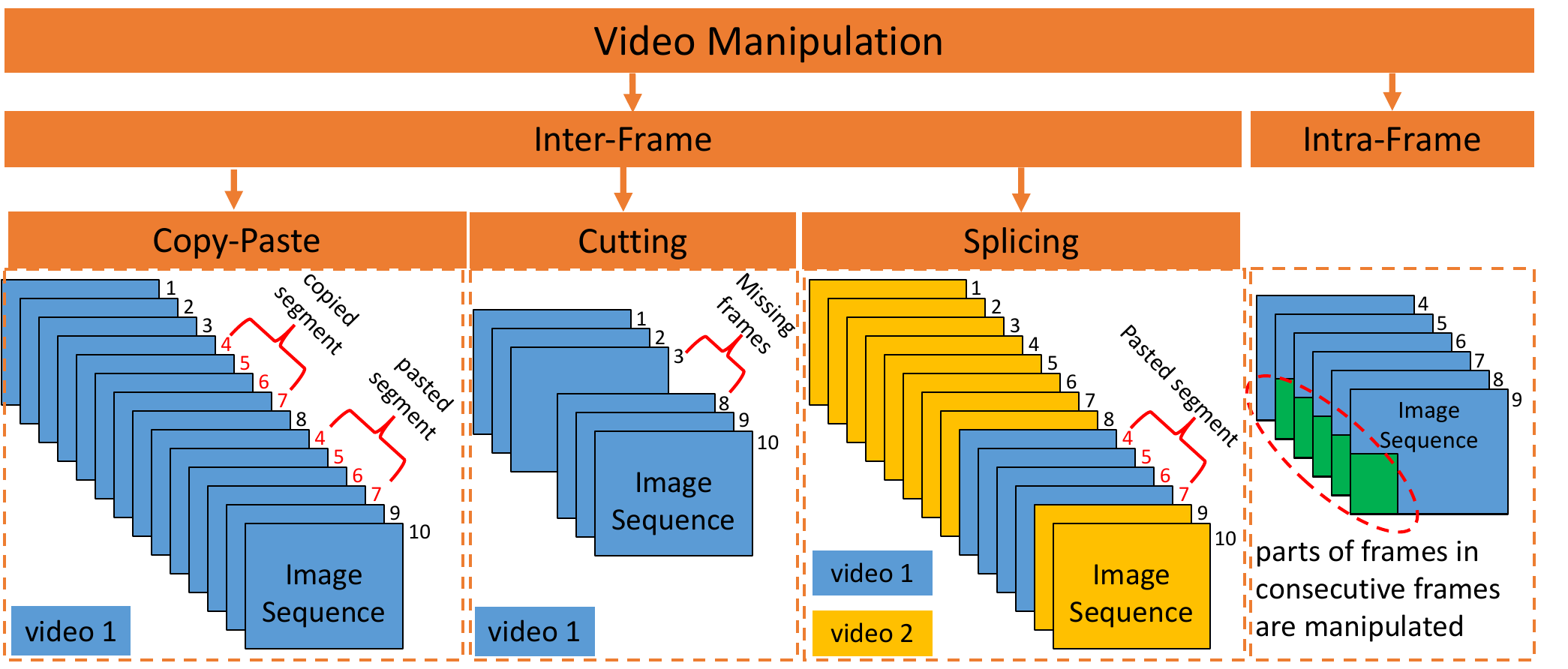}
    \caption{Types and examples of video manipulation methods.}
    \label{video_forgery}
\end{figure*}

\section{Video Forensics}\label{part-3-video}

Several review papers~\cite{bestagini2012overview,javed2021comprehensive, shelke2020comprehensive, cozzolino2022multimedia, verdoliva2020media} discussed video manipulation detection techniques.  
Kaur \etal classified video manipulation as inter-frame and intra-frame~\cite{kaur2020image}. 
Inter-frame manipulation involves modifying the order of frames. 
It can be done either by splicing (inserting frames from other video sequences into an original  video sequence), cutting (deleting frames from an original video sequence, or copy-paste (duplicating frames from one temporal location to another within the original video sequence). 
Figure~\ref{video_forgery} describes these manipulation techniques further. 
Intra-frame manipulation involves modifying the contents of individual frames.
This modification is sometimes done in multiple consecutive frame sets as shown in Figure~\ref{video_forgery}. 
Javed \etal provides an in-depth review of detection methods for these manipulations~\cite{javed2021comprehensive}. 

Video signals are ``constructed'' in their acquisition and processing steps similar to other media types.
This includes the camera sensor system and model-specific in-camera processing and compression. 
Each step of this system contributes to the video fingerprints. 
Methods for manipulated video  detection often exploit the perturbations added to these video fingerprints by video editing. 
Bestagini \etal~\cite{bestagini2012overview}, Cozzolino \etal~\cite{cozzolino2022multimedia}, and Shelke \etal~\cite{shelke2020comprehensive} reviewed such detection methods in detail, which were based on camera fingerprints such as 
Photo-Response Non-Uniformity (PRNU)~\cite{bestagini2012overview,cozzolino2022multimedia} and compression artifacts. 

This overview covers only the recent works in video sequence manipulation detection.
In this paper we shall refer to video that contains synthesized content generated using deep learning methods as ``deepfake video''~\cite{guera2018deepfake, guera2019media}. 
The discussion in this section is limited to methods for video analysis that do not use any audio associated with the video sequence. 
In Section~\ref{part-6-metadata} we discuss metadata and container methods for forensic analysis in video and audio.
We will also briefly discuss analysis of both audio sequence and image stream in Section~\ref{part-8-discussion}.

\subsection{Fingerprint-Based Detection}

In~\cite{cozzolino2019extracting}, a mirror network was used to learn camera fingerprints using unaltered video sequences from the camera.
The fingerprint was used to classify the camera model that acquired the video sequence and to localize any video manipulation.
When an edited video sequence is saved, it goes through another cycle of compression, which leads to double compression artifacts that can be exploited for detecting and localizing video manipulation.
For example, Ravi \etal proposed analyzing MPEG compression artifacts using a Huber Markov Random Field (HMRF) to determine if the video was multiply compressed~\cite{ravi2014compression}. 
Subsequently, transition probability matrices of the noise were used to detect doubly compressed video sequences.
Mahfoudi \etal demonstrated that DCT coefficients had a Laplacian distribution dependent on the quantization parameter used in the encoder~\cite{mahfoudi2022statistical}, which could be used for double compression detection. 
Uddin \etal investigated HEVC/H.265~\cite{sullivan2012overview}  encoded videos and utilized both statistical and deep convolutional neural network features for multiple compression detection~\cite{uddin2022double}. 
In~\cite{kang2021edge}, the authors used Scale-Invariant Feature Transform (SIFT)~\cite{lowe1999object} features to reveal intra-frame copy-paste attacks by detecting edge lines.
In~\cite{singh2022chroma}, a difference frame was generated for each frame and the edges detected using Canny edge detector~\cite{canny1986computational} were thresholded to detect foreground manipulation.
Ouadrhiri \etal demonstrated near-duplicate video detection using content-based features such as luminescence (visual features), color motion features based on motion vectors, and high-level features~\cite{adoui2021video}.

\subsection{Deepfake Video Detection}

Deepfakes are realistic-looking synthetic video sequences generated using deep learning.
Although they may look indistinguishable from real video sequences, the deepfake video sequences featuring human faces are often inconsistent with how real human faces talk or move (\eg abnormal blinking or breathing).
To detect whether a video sequence is a deepfake, we can exploit these inconsistencies if we have access to the characteristics of unaltered video sequences. 
Several deepfake detection methods have been proposed based on this idea.

One of the earliest deepfake detection methods was proposed by G\"{u}era \etal~\cite{guera2018deepfake, guera2019media}.
The authors developed a temporal-aware pipeline to automatically detect deepfake videos. 
Their system uses a CNN to extract frame-level features.
The features are used to train a Recurrent Neural Network (RNN)~\cite{rumelhart1986learning} that learns to classify if a video has been altered or not. 

In~\cite{cozzolino2021idreveal}, the authors used temporal facial features of a person (\eg face movement) while talking.
Facial features such as shape, expression, and pose~\cite{guo2020towards} were used with a modified ResNet architecture~\cite{he2016deep,cozzolino2021idreveal}.
The network, coupled with an adversarial training strategy, provided a representation of temporal facial behaviour.
There were inconsistencies even in realistic-looking deepfakes with respect to the temporal face expression, which were used for deepfake detection. 
This method worked well on facial reenactment detection even in the presence of strong video compression. 

In~\cite{bonettini2021video}, high-level semantic facial information from an ensemble of CNNs was used to detect deepfakes. 
An attention mechanism was used to infer and focus only on the relevant parts of the face.
This mechanism along with a modified  EfficientNetB4~\cite{tan2019efficientnet} were used to infer which part of face were relevant, and the method only focused on those parts for manipulation detection.
In~\cite{montserrat2020deepfakes, hao2022deepfake}, face bounding boxes and facial landmark features from a Multi-Task Cascaded CNN (MTCNN) were used for deepfake video detection~\cite{zhang2016joint}. 
These feature were used with EfficientNetB4~\cite{tan2019efficientnet} to classify real and synthetic faces.
For each face region in the video sequence, a classification and a weight were assigned using the attention mechanism~\cite{vaswani2017attention}, which together provided the probability of the video sequence being synthetic.
For temporal features, a Recurrent Neural Network (RNN)~\cite{rumelhart1986learning} was used to merge all features and generate the final decision. 

Detecting deepfakes in highly compressed or low quality video sequences is challenging.
Guhagarkar \etal used super resolution to improve the quality of the video sequence, which made it easier for the CNN to detect features for each frame~\cite{guhagarkar2021novel}.
These features were then used by a Long Short-Term Memory (LSTM) network~\cite{hochreiter1997long} to capture temporal features for deepfake video detection.  

Social media platforms are often the targets of manipulated video sequences.
In~\cite{marcon2021detection}, the authors observed a performance drop of CNN-based video manipulation detection methods when they were tested on manipulated videos shared on social media.
They created a dataset of manipulated videos (non-shared), shared them on a social platform, and then downloaded them (shared-videos).
For both non-shared and shared videos, several CNN networks were trained.
The results showed that fine-tuning networks trained on non-shared videos for shared videos helped in improving performance. 

Often, CNN-based deepfake video detection methods fail to generalize on datasets that were not used during training.
For example, Bondi \etal analyzed cross-dataset performance of EfficientNetB4~\cite{bondi2020training}.
The analysis showed that in limited availability of datasets, triplet loss provided good intra-dataset and cross-dataset performance and was well-suited for generalization.
For large datasets, augmenting data and using Binary Cross Entropy (BCE)-trained CNN architectures gave good results.

CNN-based architectures have received great attention in deepfake video detection.
With improvements in these methods, the size, memory, and computational requirement of the networks are growing.
Deeper networks also require more training data.
To overcome these challenges, Hinton \etal proposed Capsule network~\cite{hinton2011transforming}. The network has three blocks: a CNN network, a primary capsule block having convolutional layers with a statistical pooling layer, and an output capsule block. Nguyen \etal proposed a method using the Capsule network for deepfake video detection~\cite{nguyen2022capsuleforensics}. 
There are many primary capsules in the network and each captures a separate artifact, which are fused using a dynamic routing method. 
The output block produces a probability of the frame in the video sequence being synthetic or real.
The features learned by this network are easy to visualize and the network performed well on even unseen manipulations. 
Mazzia \etal and Huang \etal worked on making these networks more efficient by reducing the number of parameters and modifying the attention mechanism, respectively~\cite{mazzia2021efficientcapsnet, huang2020dacapsnet}. 

Some methods also focus on detecting deepfake videos not featuring human faces such as GAN-synthesized street videos.
In~\cite{alamayreh2021detection}, frames from the video sequence were used as an input to an Xception CNN. The binary classification of each frame being real or synthetic was then aggregated for final decision.
\section{Audio Forensics}\label{part-4-audio}

In this section we examine work in audio forensics. 
We will use the term ``audio'' to indicate any type of acoustic signal.
Manipulation detection methods are usually classified into two categories: container-based methods and content-based methods~\cite{zakariah2017digital, bevinamarad2020audio}.
Content-based methods involve analysis done only on the content (\eg temporal or frequency description) and information derived from it. 
In contrast, container-based methods analyze audio container metadata and file structures for audio manipulation detection. 
Some attacks also manipulate metadata and file structures, which affect the performance of container-based methods.
Our discussion in this section is limited to content-based approaches.
In Section~\ref{part-6-metadata} we discuss metadata and container methods for forensic analysis in video and audio.
We have divided the content-based methods in this section into methods for manipulated audio signal detection and methods for synthetic speech detection.

\subsection{Methods for Manipulated Audio}

One type of audio manipulation involves removing parts of an audio signal or copying them to another location within the same signal. 
Another type of manipulation involves pasting parts of an audio signal into another signal to create a spliced audio signal~\cite{bevinamarad2020audio}. 
From acquisition to compression, each block in a digital audio system leaves a fingerprint, which can be analyzed for manipulation detection.
For example, the microphone used for audio acquisition leaves characteristic fingerprints which are detectable in the frequency domain~\cite{buchholz2009microphone} and background noise~\cite{ikram2010digital}. 
These fingerprints can be used to attribute the audio signal to the microphone device. 
The existence of signals from more than one microphone can be an indication of audio signal tampering~\cite{cuccovillo2013audio}.
The environment in which the audio signal is recorded is referred to as the acoustic environment. 
The acoustic environment also contributes to the audio as signature smearing and ambient noise~\cite{gupta2011current, zakariah2017digital}.
The presence of multiple dissimilar acoustic environments within an audio signal can also be used for audio splicing detection~\cite{gupta2011current, zakariah2017digital}.
In~\cite{zhao2014audio}, magnitudes of channel impulse response were captured using the audio spectrum and were used to classify the acoustic environment for small temporal segments of the audio signal.

After manipulation, the audio signal is typically re-compressed~\cite{zakariah2017digital}. In~\cite{yang2010detecting,liu2010detection,luo2016detection}, the authors explored the detection of double compression within an audio signal for manipulation detection.
Such detection techniques often exploit features from Modified Discrete Cosine Transform (MDCT) coefficients used during compression~\cite{iso-mp3,brandenburg1999mp3}. 
In~\cite{xiang2022forensic}, authors used MDCT coefficients with other MP3 codec data such as scalefactors, quantization step sizes, Huffman table indices, and sub-band window selection information to train a transformer to identify temporal location of multiply compressed audio signal.

To detect a copy-move attack in a speech signal, the signal can be divided into several voice segments and features such as pitch~\cite{yan2019robust}, Mel Frequency Cepstral Coefficients (MFCC)~\cite{akdeniz2021detection}, and Delta-MFCC~\cite{akdeniz2021detection} can be used.
Higher similarity (\eg similarity evaluated using Pearson correlation coefficient)~\cite{yan2019robust} between features of two segments can indicate a copy-move attack.


\subsection{Methods for Synthesized Audio}

Recent deep-learning based speech synthesis and voice conversion systems~\cite{kim2021conditional, wang2021prosody, popov2021gradtts, zhou2021seen} can generate realistic human speech, which can fool Automatic Speaker Verification (ASV) systems.
This is referred to as Logical Access (LA) attack to speech~\cite{yamagishi2021asvspoof}. 
Synthetic audio can also be uploaded to online platforms to spread misinformation.
Such attacks are referred to as deepfake attacks in~\cite{yi2022add}. 
Challenges such as ASVspoof2021~\cite{yamagishi2021asvspoof} encourage research in detecting such deepfake audio signals and LA attacks. When a recorded audio signal is replayed to fool ASV systems, the attacks are referred to as physical access attacks~\cite{yamagishi2021asvspoof}. We have categorized synthetic audio detection methods into three categories: feature-based, image-based, and waveform-based. 

\begin{figure}[htpb]
    \centering
    \includegraphics[width=0.4\textwidth]{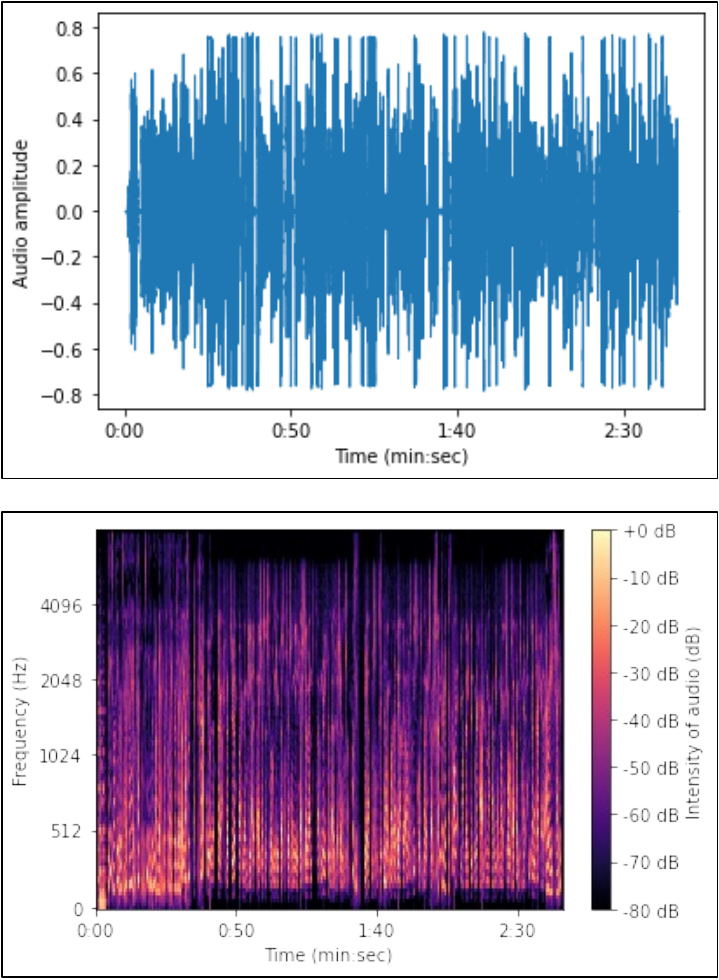}
    \caption{Example of an audio signal and its mel-spectrogram.}
    \label{audio_mel}
\end{figure}

\paragraph{Feature-Based Approaches:}
Audio features either from short-term window transforms such as Mel Frequency Cepstral Coefficients (MFCC)~\cite{akdeniz2021detection} or from long-term window transforms such as Constant-Q Cepstral Coefficients (CQCC)~\cite{todisco2017constant} are used to detect audio attacks. 
Hassan \etal used short-term spectral features consisting of MFCC, Gammatone Cepstral Coefficient (GTCC), spectral flux, and spectral centroid~\cite{hassan2021voice} to detect LA attacks. 
These features were used with a Recurrent Neural Network (RNN)~\cite{rumelhart1986learning} for detection. 
In~\cite{subramani2020learning}, a normalized log-spectrogram of temporal segments was used for classifying the signal as synthetic or real. 
For classifying each temporal segment, an Efficient Convolutional Neural Network (EfficientCNN)~\cite{tan2019efficientnet} and its residual variant RES-EfficientCNN were used.
In~\cite{yang2020longterm}, the authors investigated long-term features based on Constant-Q Transform (CQT).
Das \etal and Li \etal proposed long-term and high-frequency features such as Inverted Constant-Q Coefficient (ICQC), Inverted Constant-Q Cepstral Coefficient (ICQCC), and Long-term Variable Q Transform (L-VQT)~\cite{das2019long,li2022longterm} for detection.
These features were then passed through a deep neural network to detect synthetic audio signals.
In~\cite{li2021replay}, the authors trained a Res2Net network with features such as log power magnitude spectrogram, Linear Frequency Cepstral Coefficient (LFCC), and CQT. 
Among all these features, the CQT-trained Res2Net performed better in detecting LA attacks. 
Various features used for synthetic speech detection have been reviewed in detail in~\cite{mittal2021automatic}.

\paragraph{Image-Based Approaches:}
In image-based synthetic audio detection, the spectrogram or the melspectrogram of the audio data is treated as an image and then analyzed using computer vision methods. 
A melspectrogram refers to a spectrogram with frequencies in mel scale~\cite{choi2021comparison,mel}.
It graphically represents variation of frequency and intensity with respect to time.
Figure~\ref{audio_mel} shows a melspectrogram. 
Bartusiak \etal used normalized gray-scale spectrograms of audio signal for synthetic speech detection using a CNN and a convolution transformer~\cite{bartusiak2021frequency, bartusiak2021synthesized, hao2022deepfake}. 
While in~\cite{khochare2021deep}, the authors trained a temporal CNN and a spatial transform network using melspectrograms. 
For synthetic audio detection, these image-based methods outperformed feature-based methods including the ones using features related to energy, bandwidth, frequency, and short-term transform features such as MFCCs. 

\paragraph{Waveform-Based Approaches:}
In waveform-based methods, the audio signal or waveform is used as an input to a deep neural network. The authors of~\cite{hua2021towards} hypothesized that deep networks learned high-level features while spoofing generated subtle artifacts that could be better captured by shallower networks. They proposed the Time-domain Synthetic Speech Detection Network (TSSDNet) having multiple blocks similar to ResNet~\cite{he2016deep} and Inception~\cite{szegedy2015going}. 
In~\cite{chintha2020recurrent}, the authors presented a convolutional RNN to detect synthetic speech. 
In this method, features from CNN were passed through a bidirectional LSTM~\cite{hochreiter1997long} for synthetic speech detection. 

In summary, existing methods focus on detecting synthetic audio to counter LA and deepfake attacks. More challenges (\eg Audio Deep Synthesis Detection Challenge 2022~\cite{yi2022add}) will promote research further in this area.

\section{Text and Document Forensics}\label{part-5-documents}

Documents such as news articles, bank receipts, scientific publications, social media posts, and business forms can be manipulated with malicious intent. 
Manipulated news articles can spread fake news or misinformation to negatively influence public opinion.
Often, the documents circulated in companies do not contain marks to check their integrity visually. 
In such cases, document manipulation can lead to financial losses.
Document manipulation detection methods have been proposed to prevent such scenarios~\cite{doermann2010evolution}. 
Most common document manipulation detection methods can be of two types: (1) attributing the document to its original printer/scanner to check if it is authentic and not synthesized~\cite{elkasrawi2014printer,khanna2007scanner}; and (2) checking for irregularities in a document by detecting manipulation~\cite{jawahar2020automatic}. 

We categorize document manipulation detection analysis as text forensics analysis and document forensics analysis. 
We also briefly discuss multimodal analysis involved in documents.

\paragraph{Text Forensics Analysis:}
Text forensics methods detect inconsistencies by analyzing the text of the document. 
Several methods have been proposed to detect text manipulation using machine learning. 
Aldwairi \etal proposed a method to identify potential fake news in websites by detecting misleading words and informal phrases in the website links, analyzing the length of titles and the use of punctuation marks~\cite{aldwairi2018detecting}. 
The method used text to detect malicious websites after they were retrieved by the search engine and notified the user that they may contain misinformation. 
Ahmad \etal used linguistic features with an ensemble of different text classifiers to distinguish fake news articles from real ones~\cite{ahmad2020fake}. 
These linguistic features included punctuation, emotions associated with words (positive or negative), and grammar. 
Recent Text Generative Models (TGMs)~\cite{bakhtin2021residual,gpt2,gpt3,bert2019} can produce convincing human language-like text which can spread fake news, generate false online reviews on products, and can be used for spamming emails. 
Zellers \etal proposed Grover, which is both a text generator and detector for fake news~\cite{zellers2019defending}. 
Given an article heading, Grover generates text for the article and can produce persuasive misinformation. It can also discriminate fake content using artifacts introduced by the Grover generator.
Jawahar \etal discussed recent methods for detecting text generated using TGMs in detail~\cite{jawahar2020automatic}. 

Author attribution and author verification are also important and challenging problems in text forensics analysis.
Author attribution relates to identifying the author of a given document to prevent deceit, while author verification is concerned with finding out whether multiple given documents were written by the same author.
Traditionally, methods for authorship analysis are based on the extraction of stylometric features.
Stylometric features, \ie statistical features of a document, can be divided into several categories, such as lexical features, character features, syntactic features, structural features, and semantic features~\cite{stamatatos2009survey,ding2017learning, bhargava2013stylometric,kalgutkar2019code}.
Halvani \etal examined 12 existing author verification methods on their own self-compiled corpora, where each corpus focuses on a different aspect of applicability~\cite{halvani2019assessing}.
Driven by the popularity and advancement of deep learning in natural language processing, researchers have integrated deep learning models into the feature extraction task for authorship attribution and verification~\cite{fabien2020bertaa, shrestha2017convolutional}.

\paragraph{Document Forensics Analysis:}
Document image forensics methods treat a document as an image and then use variations of image forensics methods to analyze the document. 
Images of handwritten signatures can be used for verification of printed documents. 
Gideon \etal proposed a method using Convolutional Neural Networks (CNNs) to detect the probability of a handwritten signature being genuine or manipulated~\cite{gideon2018handwritten}.
Beusekom \etal analyzed text-line alignment in document images for their authentication~\cite{beusekom2012text}. 
They detected text-lines in the document image and measured skew angles between lines to determine the probability of the document being manipulated. 
Cruz \etal used uniform Local Binary Patterns (LBP) for texture features that are characteristic of manipulated regions~\cite{cruz2017local}.
They used a Support Vector Machine (SVM)~\cite{noble2006what} to identify whether patches of document images were manipulated. 
Document images can also be checked for manipulation by attributing them to an original scanner or printer, which leave their fingerprints on the images~\cite{rocha2017,chiang2009printer}.


The prevalence of scientific publications with problematic images has risen significantly over the past decade~\cite{Bik2016prevalence}.
Retouched or reused images are important reasons for retractions of scientific publications.
These edits are examples of scientific misconduct that undermine the integrity of the presented research.
Traditionally, progress in scientific publishing has relied on a relatively slow cycle of peer review, where human experts need to inspect the authenticity of images~\cite{lonni2022correction, daniel2022sila}.
Technologies that can automatically examine figures in published scientific papers will benefit experts who need to accomplish such an important and difficult task.
Zhuang \etal analyzed graphical integrity issues in open access publications by verifying if the size of shaded areas in scientific figures were consistent with their corresponding quantities~\cite{zhuang2021graphical}.
Moreira \etal demonstrated a human-in-the-loop end-to-end scientific publication analysis process. 
It starts by extracting content from uploaded PDFs, performs image manipulation detection on the automatically extracted figures, and ends with image provenance graphs expressing the relationships between the images in question, allowing experts to examine potential problems~\cite{daniel2022sila}.

\paragraph{Multimodal Analysis:}
Document forensics often overlaps with multimodal analysis (analysis of a media which contains some combination of image, video, audio, and text). 
For example, news articles frequently contain images, captions, and text. 
Several methods have been proposed for fact-checking news articles.
Vo \etal proposed a method for fact-checking articles using both text and images through a Multimodal Attention Network (MAN)~\cite{vo2020where}.
Fung \etal used a graph-based neural network for fake news detection~\cite{fung2021infosurgeon}. 

In articles with multimedia assets, authors may manipulate images to match text or manipulate the text in article to convince readers of what they are trying to convey, usually to present a false narrative. 
Some authors may even misuse some images in their articles in hopes of grabbing attention.
Often, such manipulations lead to image-text mismatch, which can be used for cross-model manipulation detection.
Image-text matching~\cite{lee2018stacked,xu2020crossmodal} refers to checking semantic similarity of an image in an article with its caption.
Zhang \etal and Li \etal proposed a projection matching loss~\cite{zhang2018deep} and a Visual Semantic Reasoning Network~\cite{li2019visual}, respectively, for image-text matching.
Li \etal proposed a deep learning method known as Object-Semantics Aligned Pre-training (Oscar), which used object tags detected in images as anchor points to significantly ease the learning of alignments~\cite{li2020oscar}.

Tan \etal used visual-semantic representations for detecting inconsistencies in news articles~\cite{tan2020detecting}. 
These visual-semantic representations which included representations of text, objects in the images, and captions were used for classifying the article as either machine-generated or human-generated.
McCrae \etal proposed a fusion method to detect manipulations in social media posts by identifying inconsistencies between videos and their captions~\cite{mccrae2021multi}.
These methods used machine learning and deep learning methods to classify media as genuine or manipulated.

\section{Metadata Forensics}\label{part-6-metadata}

\begin{figure}[ht]
    \centering
    \usetikzlibrary{positioning, fit, arrows.meta, calc}

\definecolor{md-node-c3}{HTML}{55868C}
\definecolor{md-node-c1}{HTML}{CACAAA}
\definecolor{md-node-c2}{HTML}{C8AB83}


\tikzset{
    bnd/.style={
        rotate=90,
        text width=6em,
        fill=md-node-c1,
        text centered,
        minimum height=1.5em
    },
    tnd/.style={
        fill=md-node-c1,
        text centered,
        minimum height=1.5em
    },
    mnd/.style={
        rotate=90,
        text width=1.5em,
        text centered,
        font=\fontsize{9}{8}\selectfont,
        outer sep=0.1em
    }
}

\gdef\MakeMetadataFigureA{

    \begin{tikzpicture}
        
        \begin{scope}
            \node[bnd, fill=md-node-c2] (container-1) {\texttt{ffprobe}};
            \node[bnd, right=of container-1.east, anchor=east, yshift=-2em, , fill=md-node-c2] (container-2) {MP4};
            \node[fit=(container-1)(container-2)] (container-bbox) {};
        \end{scope}

        \begin{scope}
            \node[bnd, right=of container-2.east, anchor=east, yshift=-2em, fill=md-node-c2] (encp-1) {H.264};
            \node[bnd, right=of encp-1.east, anchor=east, fill=md-node-c2] (encp-2) {H.265};
            \node[bnd, right=of encp-2.east, anchor=east, fill=md-node-c3] (encp-3) {MP3};
            \node[bnd, right=of encp-3.east, anchor=east, fill=md-node-c3] (encp-4) {AAC};
            \node[fit=(encp-1)(encp-4)] (encp-bbox) {};
        \end{scope}

        \path
            let
            \p1=(container-bbox.east),\p2=(container-bbox.west),
            \n1={3em+\x1-\x2-2*\pgfkeysvalueof{/pgf/inner xsep}-\pgflinewidth}
            in
            node
            [tnd,text width=\n1]
            (n-container)
            at ($(container-bbox.north)+(0em, 2em)$) {Container Metadata};

        \path
            let
            \p1=(encp-bbox.east),\p2=(encp-bbox.west),
            \n1={0.5em+\x1-\x2-2*\pgfkeysvalueof{/pgf/inner xsep}-\pgflinewidth}
            in
            node
            [tnd,text width=\n1]
            (n-encp)
            at ($(encp-bbox.north)+(0em, 2em)$) {Encoding Parameters};

        
        \draw (container-1.east) -- (container-1.east|-n-container.south);
        \draw (container-2.east) -- (container-2.east|-n-container.south);
        
        \draw (encp-1.east) -- (encp-1.east|-n-encp.south);
        \draw (encp-2.east) -- (encp-2.east|-n-encp.south);
        \draw (encp-3.east) -- (encp-3.east|-n-encp.south);
        \draw (encp-4.east) -- (encp-4.east|-n-encp.south);
        
        \node[fit=(n-container)(n-encp), inner sep=0pt, outer sep=0pt] (all-metadata) {};
        \path
            let
            \p1=(all-metadata.east),\p2=(all-metadata.west),
            \n1={\x1-\x2-2*\pgfkeysvalueof{/pgf/inner xsep}-\pgflinewidth}
            in
            node
            [tnd,text width=\n1, above=of all-metadata, yshift=-1.5em]
            (media-forensics)
            {Metadata Media Forensics};
        
        \draw (media-forensics.south-|n-container)--(n-container);
        \draw (media-forensics.south-|n-encp)--(n-encp);
        
    \begin{scope}[node distance=1.5em and 1.5em]
        \node[mnd,below=of container-1.west, anchor=east] (ana-1) {\ding[1.5]{192}};
        \node[mnd,below=of container-2.west, anchor=east] (ana-2) {\ding[1.5]{192}\ding[1.5]{194}\\ \ding[1.5]{195}\ding[1.5]{196}};
        \node[mnd,below=of encp-1.west, anchor=east] (ana-3) {\ding[1.5]{193}\ding[1.5]{194}};
        \node[mnd,below=of encp-2.west, anchor=east] (ana-4) {\ding[1.5]{193}};
        \node[mnd,below=of encp-3.west, anchor=east] (ana-5) {\ding[1.5]{193}};
        \node[mnd,below=of encp-4.west, anchor=east] (ana-6) {\ding[1.5]{193}};
    \end{scope}

    \draw[dotted] (container-1.west)--(ana-1.east);
    \draw[dotted] (container-2.west)--(ana-2.east);
    \draw[dotted] (encp-1.west)--(ana-3.east);
    \draw[dotted] (encp-2.west)--(ana-4.east);
    \draw[dotted] (encp-3.west)--(ana-5.east);
    \draw[dotted] (encp-4.west)--(ana-6.east);
    
    \coordinate (fig-hcenter) at ($0.5*(n-container.west)+0.5*(n-encp.east)$);
    
    \coordinate (sep-line1-left) at ($(n-container.west|-ana-1.south)+(0em, -1.5em)$);
    \coordinate (sep-line1-right) at ($(n-encp.east|-ana-1.south)+(0em, -1.5em)$);
    
    \draw[draw=gray] (sep-line1-left)--(sep-line1-right);
        
    \begin{scope}[shift={($(sep-line1-left)+(0.6em, -0.7em)$)}, name prefix=g1-, node distance=0.5em and 1em]
        \node[minimum width=1em, minimum height=1em, fill=md-node-c2] (nl1) {};
        \node[font=\small, anchor=west] (nr1) at ($(nl1.east)+(0.1em, 0em)$) {Video Method};
        
        \node[minimum width=1em, minimum height=1em, fill=md-node-c3, right=of nr1] (nl2) {};
        \node[font=\small, anchor=west] (nr2) at ($(nl2.east)+(0.1em, 0em)$) {Audio Method};
        
        \coordinate (nl3-l) at ($(nr2.east)+(0.6em, 0em)$);
        \coordinate (nl3-r) at ($(nl3-l)+(1em, 0em)$);
        \draw[dotted] (nl3-l)--(nl3-r);
        \node[font=\small, anchor=west] (nr3) at ($(nl3-r.east)+(0.1em, 0em)$) {Forensic Task};
    \end{scope}
    
    \coordinate (sep-line2-left) at ($(sep-line1-left|-g1-nl1.south)+(0em, -0.2em)$);
    
    \coordinate (sep-line2-right) at (sep-line2-left-|sep-line1-right);
    
    \draw[draw=gray] (sep-line2-left)--(sep-line2-right);
    
    \coordinate (tab-center) at ($0.5*(sep-line2-left)+0.5*(sep-line2-right)+(0em,-2em)$);
    
    \node[font=\normalsize] at (tab-center) {
        \begin{tabular}{m{0.6\linewidth}m{0.6\linewidth}}
            \\
            \\
            \\
            \ding[1.3]{192}~Manipulation Detection \\ \ding[1.3]{193}~Double Compression Detection \\
            \ding[1.3]{194}~Source Device Attribution \\
            \ding[1.3]{195}~Social Media Analysis\\
            \ding[1.3]{196}~Manipulation Tool Attribution
        \end{tabular}
    };
        
    \end{tikzpicture}

}
    \MakeMetadataFigureA
    \caption{Metadata media forensic analysis types and their corresponding target forensic tasks described in Section \ref{part-6-metadata}.}
    \label{fig:metadata_figure}
\end{figure}
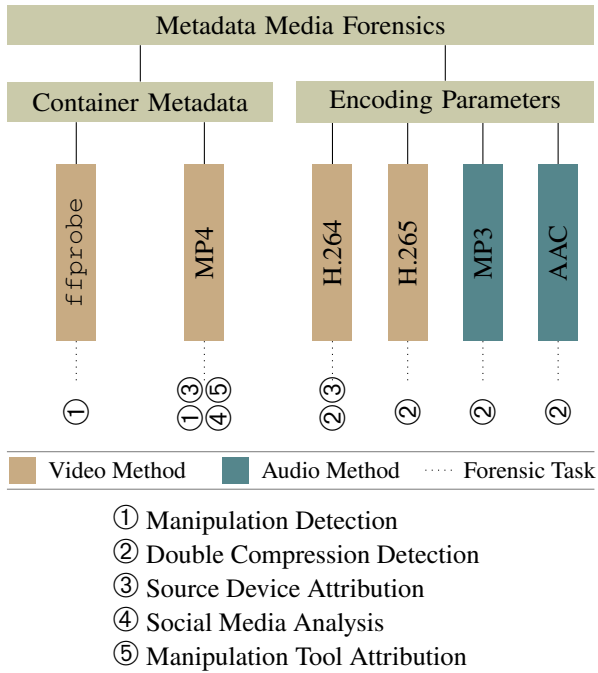

Almost all media files contain metadata.
For example, the MP4 video container stores information about the modification date, video bitstream format, and video length~\cite{iuliani2019video}.
MP3 compressed audio signals contain encoding parameters about the details of windowing, quantization, and Huffman encoding~\cite{raissi2002theory}.
Metadata may not be used by end users directly, but media files cannot be decoded correctly without them.
Many recent publications have shown that metadata in video and audio files can be used for forensic analysis~\cite{guera2019we,guera2019media, iuliani2019video,vazquez2020video,yang2010detecting,xiang2022forensic}. 
There are mainly two advantages for forensic analysis using metadata.
First, since the metadata format and structure are strictly specified in the standards of video/audio bitstream, it is more challenging for one to conceal or alter the forensic traces left in metadata without damaging the media file. 
Second, many existing video/audio manipulation or synthesis methods do not attempt to hide metadata traces, which increases the chances of detection.
In this section, we examine video and audio forensics methods using metadata.
In Figure \ref{fig:metadata_figure}, we show the categorization and target forensic scenarios for metadata media forensics techniques described in this section.


The metadata stored in video containers can be used for forensic analysis.
G\"{u}era \etal used metadata extracted by the \texttt{ffprobe}\footnote{\url{https://ffmpeg.org/ffprobe.html}} tool for video manipulation detection~\cite{guera2019we, guera2019media}.
More metadata and forensic traces can be retrieved by in-depth analysis of a specific video container format.
For example, MP4 is one of the most popular video containers~\cite{iso-mp4file}.
Iuliani \etal exploited the tree structure of MP4 containers for video manipulation detection and video source attribution using an unsupervised probabilistic framework~\cite{iuliani2019video}.
In \cite{yang2020efficient}, a similar MP4 metadata feature processing method was used together with decision tree classifiers for forensic analysis. 
In addition to video forensics tasks analyzed in \cite{iuliani2019video}, the authors also considered video manipulation tool attribution and the influence of social media in metadata-based video forensics.
This metadata feature processing method is extended in \cite{xiang2021forensic}, where the authors improved the quality of metadata features by parsing more metadata, using type-specific feature construction strategies, and reducing the size of feature vectors. 

The encoding parameters in video bitstream can be used for forensic analysis.
In \cite{su2009asource}, the motion vectors from motion compensation were used for video device attribution.
The encoding parameters are also used extensively for double compression detection. 
In \cite{yao2020double,vazquez2020video,mahfoudi2022statistical,uddin2022double}, the authors investigated double compression detection using encoding parameters in H.264 bitstreams.
In \cite{fang2019detection,he2021framewise,jiang2020detection,uddin2022double}, the authors used encoding parameters in H.265 video bitstreams for double compression detection.
Altinisik \etal used both encoding parameters and video container metadata for video device attribution, which resulted in improved performance compared to methods using container metadata only~\cite{altinisik2022camera}.

The encoding parameters in audio bitstream can be used for forensic analysis.
Most related work focus on two popular audio compression methods, namely MP3 and AAC~\cite{brandenburg1999mp3}.
In \cite{liu2010detection,yang2010detecting,qiao2013improved}, the authors used the Modified Discrete Cosine Transform (MDCT) coefficients for double compression detection in MP3.
Ma \etal used the MP3 scalefactors for double compression detection~\cite{ma2014detecting}.
In \cite{ma2014huffman}, the authors achieved MP3 double compression detection using the Huffman table indices.
Yan \etal used scalefactors and Huffman table indices information in MP3 files for double and triple compression detection~\cite{yan2018compression}.
In \cite{bianchi2014detection}, the authors used the MDCT coefficients in MP3 compressed signals for multiple compression localization.
Xiang \etal used a transformer neural network to process many types of MP3 encoding parameters, and their method achieved multiple compression localization with high accuracy~\cite{xiang2022forensic}.
In \cite{jin2016efficient}, the authors used the Huffman table indices in AAC files for double compression detection.
Huang \etal  achieved AAC double compression detection using the Quantized MDCT (QMDCT) coefficients \cite{huang2018aac}.
In \cite{huang2018aac2}, the authors used the scalefactors from AAC compressed signals for double compression detection.

\section{Discussion and Future Work}\label{part-8-discussion}

Research in media forensics is advancing rapidly, driven by competition with advancements in techniques that manipulate and generate media.
In this paper we discuss some of the recent research for detecting manipulated images, videos, audio signals, and documents.
Developments in deep learning have made it easy to individually generate deepfake audio, image, video, and text.
Multimedia assets typically have more than one media type present (\eg a video has image sequences and audio sequences, news articles have text along with images).
Current counter-forensic approaches struggle to simultaneously manipulate these multiple data modalities consistently.
Future work should focus on leveraging multi-modal analysis to detect these inconsistencies in order to identify manipulated media.

For example, in \cite{hosler2021do}, the authors detected deepfake videos of humans by analyzing both audio and image sequences individually by looking for emotional inconsistencies across them. 
Audio and image sequences were divided into temporal segments.
For each audio segment using speech features, an LSTM was used to predict the subject's emotion.
For each image sequence segment, a similar LSTM using facial features was also used to predict emotions.
The Lin’s Concordance Correlation Coefficient~\cite{hosler2021do} estimated the correlation between the video and audio and predicted inconsistencies of the emotions.
We aim to apply a similar concept to image-text cross-modal forensic analysis, with an approach that exploits the alignment between object labels and text, as well as attention regions in an image (such as faces) and name entities in text to examine the consistency between image and text.

Another direction to explore is combining statistical features and neural network features together to improve attribution and verification of a document.
Lastly, we can look at intent prediction.
Multiple media assets may be manipulated for a common purpose.
Identifying common manipulation motivations of different media types would provide forensics analysts with more information about the nature of the manipulations. 

\section*{Acknowledgments}
This paper is based on research sponsored by the Defense Advanced Research Projects Agency (DARPA) and the Air Force Research Laboratory (AFRL) under agreement numbers FA8750-20-2-1004 and FA8750-16-2-0173. The U.S. Government is authorized to reproduce and distribute reprints for Governmental purposes notwithstanding any copyright notation thereon. The views and conclusions contained herein are those of the authors and should not be interpreted as necessarily representing the official policies or endorsements, either expressed or implied, of DARPA, AFRL, or the U.S. Government.

Address all correspondence to Edward J. Delp at \url{ace@ecn.purdue.edu}. 

{
\fontsize{9}{9}\selectfont
\let\oldthebibliography\thebibliography
\let\endoldthebibliography\endthebibliography
\renewenvironment{thebibliography}[1]{
  \begin{oldthebibliography}{#1}
    \setlength{\itemsep}{0em}
    \setlength{\parskip}{0em}
}
{
  \end{oldthebibliography}
}
\bibliographystyle{ieeetran}
\bibliography{refs}

\begin{thebibliography}{100}
\providecommand{\url}[1]{#1}
\csname url@samestyle\endcsname
\providecommand{\newblock}{\relax}
\providecommand{\bibinfo}[2]{#2}
\providecommand{\BIBentrySTDinterwordspacing}{\spaceskip=0pt\relax}
\providecommand{\BIBentryALTinterwordstretchfactor}{4}
\providecommand{\BIBentryALTinterwordspacing}{\spaceskip=\fontdimen2\font plus
\BIBentryALTinterwordstretchfactor\fontdimen3\font minus
  \fontdimen4\font\relax}
\providecommand{\BIBforeignlanguage}[2]{{%
\expandafter\ifx\csname l@#1\endcsname\relax
\typeout{** WARNING: IEEEtran.bst: No hyphenation pattern has been}%
\typeout{** loaded for the language `#1'. Using the pattern for}%
\typeout{** the default language instead.}%
\else
\language=\csname l@#1\endcsname
\fi
#2}}
\providecommand{\BIBdecl}{\relax}
\BIBdecl

\bibitem{gimp}
``{GIMP - GNU Image Manipulation Program},'' \url{https://www.gimp.org/}, 2021.

\bibitem{goodfellow2016deep}
I.~Goodfellow, Y.~Bengio, and A.~Courville, \emph{Deep Learning}.\hskip 1em
  plus 0.5em minus 0.4em\relax MIT Press, 2016.

\bibitem{hastie2001elements}
T.~Hastie, R.~Tibshirani, and J.~Friedman, \emph{The Elements of Statistical
  Learning: Data Mining, Inference, and Prediction}.\hskip 1em plus 0.5em minus
  0.4em\relax New York Springer, 2001.

\bibitem{perov2020deepfacelab}
I.~Perov, D.~Gao, N.~Chervoniy, K.~Liu, S.~Marangonda, C.~Um\'{e}, M.~Dpfks,
  C.~S. Facenheim, L.~RP, J.~Jiang, S.~Zhang, P.~Wu, B.~Zhou, and W.~Zhang,
  ``{DeepFaceLab: Integrated, Flexible and Extensible Face-Swapping
  Framework},'' \emph{arXiv preprint arXiv:2005.05535}, May 2020.

\bibitem{shen2018natural}
J.~Shen, R.~Pang, R.~J. Weiss, M.~Schuster, N.~Jaitly, Z.~Yang, Z.~Chen,
  Y.~Zhang, Y.~Wang, R.~Skerrv-Ryan, R.~A. Saurous, Y.~Agiomvrgiannakis, and
  Y.~Wu, ``{Natural TTS Synthesis by Conditioning Wavenet on MEL Spectrogram
  Predictions},'' \emph{Proceedings of the IEEE International Conference on
  Acoustics, Speech and Signal Processing}, pp. 4779--4783, April 2018,
  {Calgary, Canada}.

\bibitem{farid2009seeing}
H.~Farid, ``{Seeing is not Believing},'' \emph{IEEE Spectrum}, vol.~46, no.~8,
  pp. 44--51, August 2009.

\bibitem{nightingale2022aisynthesized}
S.~J. Nightingale and H.~Farid, ``{AI-Synthesized Faces are Indistinguishable
  from Real Faces and More Trustworthy},'' \emph{Proceedings of the National
  Academy of Sciences}, vol. 119, no.~8, February 2022.

\bibitem{wakefield2022deepfake}
J.~Wakefield, ``{Deepfake Presidents Used in Russia-Ukraine War},''
  \url{https://www.bbc.com/news/technology-60780142}, March 2022.

\bibitem{vaccari2021deepfakes}
C.~Vaccari and A.~Chadwick, ``{Deepfakes and Disinformation: Exploring the
  Impact of Synthetic Political Video on Deception, Uncertainty, and Trust in
  News},'' \emph{Social Media + Society}, vol.~6, no.~1, pp. 1--12, February
  2020.

\bibitem{lyu2021fighting}
S.~Lyu, ``{Fighting AI-Synthesized Fake Media},'' \emph{Proceedings of the
  Workshop on Synthetic Multimedia - Audiovisual Deepfake Generation and
  Detection}, pp. 1--2, October 2021.

\bibitem{battiato2016multimedia}
S.~Battiato, O.~Giudice, and A.~Paratore, ``{Multimedia Forensics: Discovering
  the History of Multimedia Contents},'' \emph{Proceedings of the International
  Conference on Computer Systems and Technologies}, pp. 5--16, June 2016,
  {Palermo, Italy}.

\bibitem{piva2013overview}
A.~Piva, ``{An Overview on Image Forensics},'' \emph{International Scholarly
  Research Notices}, vol. 2013, pp. 1--22, January 2013.

\bibitem{nguyen2013counterforensics}
R.~Boehme and M.~Kirchner, ``{Counter-forensics: Attacking Image Forensics},''
  in \emph{{Digital Image Forensics: There is More to a Picture than Meets the
  Eye}}.\hskip 1em plus 0.5em minus 0.4em\relax Springer, July 2013, pp.
  327--366.

\bibitem{farid2009image}
H.~Farid, ``{Image Forgery Detection},'' \emph{IEEE Signal Processing
  Magazine}, vol.~26, no.~2, pp. 16--25, March 2009.

\bibitem{delp2005multimedia}
E.~J. Delp, ``{Multimedia Security: The 22nd Century Approach},''
  \emph{Multimedia Systems}, vol.~11, no.~2, pp. 95--97, December 2005.

\bibitem{chen2008determining}
M.~Chen, J.~Fridrich, M.~Goljan, and J.~Lukas, ``{Determining Image Origin and
  Integrity Using Sensor Noise},'' \emph{IEEE Transactions on Information
  Forensics and Security}, vol.~3, no.~1, pp. 74--90, February 2008.

\bibitem{khanna2007scanner}
N.~Khanna, A.~K. Mikkilineni, G.~T.~C. Chiu, J.~P. Allebach, and E.~J. Delp,
  ``{Scanner Identification Using Sensor Pattern Noise},'' \emph{Proceedings of
  the Society of Photo-Optical Instrumentation Engineers}, vol. 6505, pp.
  563--573, March 2007, {Florence, Italy}.

\bibitem{shao2020forensic}
R.~Shao and E.~J. Delp, ``{Forensic Scanner Identification Using Machine
  Learning},'' \emph{Proceedings of the IEEE Southwest Symposium on Image
  Analysis and Interpretation}, pp. 1--4, March 2020, {Virtual}.

\bibitem{khanna2007sensor}
N.~Khanna, A.~K. Mikkilineni, P.-J. Chiang, M.~V. Ortiz, S.~Suh, G.~T.-C. Chiu,
  J.~P. Allebach, and E.~J. Delp, ``{Sensor Forensics: Printers, Cameras and
  Scanners, They Never Lie},'' \emph{Proceedings of the IEEE International
  Conference on Multimedia and Expo}, pp. 20--23, July 2007, {Beijing, China}.

\bibitem{cozzolino2019extracting}
D.~Cozzolino, G.~Poggi, and L.~Verdoliva, ``{Extracting Camera-Based
  Fingerprints for Video Forensics},'' \emph{Proceedings of the IEEE/CVF
  Conference on Computer Vision and Pattern Recognition Workshop}, pp.
  130--137, June 2019, {Long Beach, CA}.

\bibitem{gui2021review}
J.~Gui, Z.~Sun, Y.~Wen, D.~Tao, and J.~Ye, ``{A Review on Generative
  Adversarial Networks: Algorithms, Theory, and Applications},'' \emph{IEEE
  Transactions on Knowledge and Data Engineering}, pp. 1--1, November 2021.

\bibitem{dhariwal2021diffusion}
P.~Dhariwal and A.~Q. Nichol, ``{Diffusion Models Beat {GAN}s on Image
  Synthesis},'' \emph{Advances in Neural Information Processing Systems},
  vol.~34, pp. 8780--8794, December 2021, {Virtual}.

\bibitem{marra2019do}
F.~Marra, D.~Gragnaniello, L.~Verdoliva, and G.~Poggi, ``{Do GANs Leave
  Artificial Fingerprints?}'' \emph{Proceedings of the IEEE Conference on
  Multimedia Information Processing and Retrieval}, pp. 506--511, March 2019,
  {San Jose, CA}.

\bibitem{gragnaniello2021are}
D.~Gragnaniello, D.~Cozzolino, F.~Marra, G.~Poggi, and L.~Verdoliva, ``{Are GAN
  Generated Images Easy to Detect? A Critical Analysis of the
  State-Of-The-Art},'' \emph{Proceedings of the IEEE International Conference
  on Multimedia and Expo}, pp. 1--6, July 2021, {Virtual}.

\bibitem{fridrich2009digital}
J.~Fridrich, ``{Digital Image Forensics},'' \emph{IEEE Signal Processing
  Magazine}, vol.~26, no.~2, pp. 26--37, March 2009.

\bibitem{yang2020survey}
P.~Yang, D.~Baracchi, R.~Ni, Y.~Zhao, F.~Argenti, and A.~Piva, ``{A Survey of
  Deep Learning-Based Source Image Forensics},'' \emph{Journal of Imaging},
  vol.~6, no.~3, March 2020.

\bibitem{bestagini2012overview}
P.~Bestagini, K.~M. Fontani, S.~Milani, M.~Barni, A.~Piva, M.~Tagliasacchi, and
  K.~S. Tubaro, ``{An Overview on Video Forensics},'' \emph{Proceedings of the
  European Signal Processing Conference}, pp. 1229--1233, August 2012,
  {Bucharest, Romania}.

\bibitem{javed2021comprehensive}
A.~R. Javed, Z.~Jalil, W.~Zehra, T.~R. Gadekallu, D.~Y. Suh, and M.~J. Piran,
  ``{A Comprehensive Survey on Digital Video Forensics: Taxonomy, Challenges,
  and Future Directions},'' \emph{Engineering Applications of Artificial
  Intelligence}, vol. 106, p. 104456, November 2021.

\bibitem{shelke2020comprehensive}
N.~A. Shelke and S.~S. Kasana, ``{A Comprehensive Survey on Passive Techniques
  for Digital Video Forgery Detection},'' \emph{Multimedia Tools and
  Applications}, vol.~80, no.~4, pp. 6247--6310, October 2020.

\bibitem{cozzolino2022multimedia}
D.~Cozzolino and L.~Verdoliva, ``{Multimedia Forensics Before the Deep Learning
  Era},'' in \emph{{Handbook of Digital Face Manipulation and Detection - From
  DeepFakes to Morphing Attacks, Series on Advances in Computer Vision and
  Pattern Recognition}}.\hskip 1em plus 0.5em minus 0.4em\relax Springer, March
  2022, vol.~1, pp. 45--67.

\bibitem{verdoliva2020media}
L.~Verdoliva, ``{Media Forensics and DeepFakes: An Overview},'' \emph{IEEE
  Journal of Selected Topics in Signal Processing}, vol.~14, no.~5, pp.
  910--932, June 2020.

\bibitem{zakariah2017digital}
M.~Zakariah, M.~K. Khan, and H.~Malik, ``{Digital Multimedia Audio Forensics:
  Past, Present and Future},'' \emph{Multimedia Tools and Applications},
  vol.~77, no.~1, pp. 1009--1040, January 2017.

\bibitem{moulin2005datahiding}
P.~Moulin and R.~Koetter, ``{Data-Hiding Codes},'' \emph{Proceedings of the
  IEEE}, vol.~93, no.~12, pp. 2083--2126, December 2005.

\bibitem{chu2016information}
X.~Chu, Y.~Chen, M.~C. Stamm, and K.~J.~R. Liu, ``{Information Theoretical
  Limit of Media Forensics: The Forensicability},'' \emph{IEEE Transactions on
  Information Forensics and Security}, vol.~11, no.~4, pp. 774--788, April
  2016.

\bibitem{kraetzer2015considerations}
C.~Kraetzer and J.~Dittmann, ``{Considerations on the Benchmarking of Media
  Forensics},'' \emph{Proceedings of the European Signal Processing
  Conference}, pp. 61--65, September 2015, {France}.

\bibitem{rocha2016authorship}
A.~Rocha, W.~J. Scheirer, C.~W. Forstall, T.~Cavalcante, A.~Theophilo, B.~Shen,
  A.~R.~B. Carvalho, and E.~Stamatatos, ``{Authorship Attribution for Social
  Media Forensics},'' \emph{IEEE Transactions on Information Forensics and
  Security}, vol.~12, no.~1, pp. 5--33, August 2016.

\bibitem{saini2016forensic}
K.~Saini and S.~Kaur, ``{Forensic Examination of Computer-Manipulated Documents
  using Image Processing Techniques},'' \emph{Egyptian Journal of Forensic
  Sciences}, vol.~6, no.~3, pp. 317--322, September 2016.

\bibitem{chiang2009printer}
P.-J. Chiang, N.~Khanna, A.~K. Mikkilineni, M.~V. Ortiz~Segovia, S.~Suh, J.~P.
  Allebach, G.~T.-C. Chiu, and E.~J. Delp, ``{Printer and scanner forensics},''
  \emph{IEEE Signal Processing Magazine}, vol.~26, no.~2, pp. 72--83, March
  2009.

\bibitem{doermann2010evolution}
D.~Doermann, ``{The Evolution of Document Authentication},'' \emph{Proceedings
  of the International Conference on Frontiers in Handwriting Recognition}, pp.
  3--3, November 2010, {Kolkata, India}.

\bibitem{bhagtani2022overview}
K.~Bhagtani, A.~K.~S. Yadav, E.~R. Bartusiak, Z.~Xiang, R.~Shao, S.~Baireddy,
  and E.~J. Delp, ``{An Overview of Recent Work in Multimedia Forensics},''
  \emph{Proceedings of the IEEE Conference on Multimedia Information Processing
  and Retrieval}, August 2022, {Virtual}.

\bibitem{nataraj2021holistic}
L.~Nataraj, M.~Goebel, T.~M. Mohammed, S.~Chandrasekaran, and B.~S. Manjunath,
  ``{Holistic Image Manipulation Detection using Pixel Co-occurrence
  Matrices},'' \emph{Electronic Imaging}, pp. 277--1--277--7, January 2021,
  {Burlingame, CA}.

\bibitem{yang2017highresolution}
C.~Yang, X.~Lu, Z.~Lin, E.~Shechtman, O.~Wang, and H.~Li, ``{High-Resolution
  Image Inpainting Using Multi-Scale Neural Patch Synthesis},''
  \emph{Proceedings of the IEEE Conference on Computer Vision and Pattern
  Recognition}, pp. 6721--6729, July 2017, {Honolulu, HI}.

\bibitem{yu2020manipulation}
I.-J. Yu, S.-H. Nam, W.~Ahn, M.-J. Kwon, and H.-K. Lee, ``{Manipulation
  Classification for JPEG Images Using Multi-Domain Features},'' \emph{IEEE
  Access}, vol.~8, pp. 210\,837--210\,854, November 2020.

\bibitem{tan2018where}
F.~Tan, C.~Bernier, B.~Cohen, V.~Ordonez, and C.~Barnes, ``{Where and Who?
  Automatic Semantic-Aware Person Composition},'' \emph{Proceedings of the IEEE
  Winter Conference on Applications of Computer Vision}, pp. 1519--1528, March
  2018, {Lake Tahoe, NV}.

\bibitem{guera2017a}
D.~G\"{u}era, Y.~Wang, L.~Bondi, P.~Bestagini, S.~Tubaro, and E.~J. Delp, ``{A
  Counter-Forensic Method for {CNN}-Based Camera Model Identification},''
  \emph{Proceedings of the IEEE Conference on Computer Vision and Pattern
  Recognition Workshops}, pp. 1840--1847, July 2017, {Honolulu, HI}.

\bibitem{bonettini2018fooling}
N.~Bonettini, L.~Bondi, D.~G\"{u}era, S.~Mandelli, P.~Bestagini, S.~Tubaro, and
  E.~J. Delp, ``{Fooling PRNU-Based Detectors Through Convolutional Neural
  Networks},'' \emph{Proceedings of the IEEE European Signal Processing
  Conference}, September 2018, {Rome, Italy}.

\bibitem{cozzolino2021spoc}
D.~Cozzolino, J.~Thies, A.~R\"{o}ssler, M.~Nie{\ss}ner, and L.~Verdoliva,
  ``{SpoC: Spoofing Camera Fingerprints},'' \emph{Proceedings of the IEEE/CVF
  Conference on Computer Vision and Pattern Recognition Workshops}, pp.
  990--1000, June 2021, {Virtual}.

\bibitem{huang2021dodging}
Y.~Huang, F.~Juefei-Xu, Q.~Guo, L.~Ma, X.~Xie, W.~Miao, Y.~Liu, and G.~Pu,
  ``{Dodging DeepFake Detection via Implicit Spatial-Domain Notch Filtering},''
  \emph{arXiv preprint arXiv:2009.09213}, November 2021.

\bibitem{zhu2017unpaired}
J.-Y. Zhu, T.~Park, P.~Isola, and A.~A. Efros, ``{Unpaired Image-to-Image
  Translation using Cycle-Consistent Adversarial Networks},'' \emph{Proceedings
  of the IEEE International Conference on Computer Vision}, pp. 2223--2232,
  October 2017, {Venice, Italy}.

\bibitem{barnes2009patchmatch}
C.~Barnes, E.~Shechtman, A.~Finkelstein, and D.~B. Goldman, ``{PatchMatch: A
  Randomized Correspondence Algorithm for Structural Image Editing},''
  \emph{ACM Transactions on Graphics}, vol.~28, no.~3, July 2009.

\bibitem{schetinger2016digital}
V.~Schetinger, M.~Iuliani, A.~Piva, and M.~M. Oliveira, ``{Digital Image
  Forensics vs. Image Composition: An Indirect Arms Race},'' \emph{arXiv
  preprint arXiv:1601.03239}, January 2016.

\bibitem{goodfellow2014generative}
I.~Goodfellow, J.~Pouget-Abadie, M.~Mirza, B.~Xu, D.~Warde-Farley, S.~Ozair,
  A.~Courville, and Y.~Bengio, ``{Generative Adversarial Nets},''
  \emph{Advances in Neural Information Processing Systems}, pp. 2672--2680,
  December 2014, {Montr\'{e}al, Canada}.

\bibitem{nazeri2019edgeconnect}
K.~Nazeri, E.~Ng, T.~Joseph, F.~Z. Qureshi, and M.~Ebrahimi, ``{EdgeConnect:
  Generative Image Inpainting with Adversarial Edge Learning},'' \emph{arXiv
  preprint arXiv:1901.00212}, January 2019.

\bibitem{karras2018progressive}
T.~Karras, T.~Aila, S.~Laine, and J.~Lehtinen, ``{Progressive Growing of GANs
  for Improved Quality, Stability, and Variation},'' \emph{Proceedings of the
  International Conference on Learning Representations}, April 2018,
  {Vancouver, Canada}.

\bibitem{karras2021stylegan}
T.~Karras, M.~Aittala, S.~Laine, E.~H\"{a}rk\"{o}nen, J.~Hellsten, J.~Lehtinen,
  and T.~Aila, ``{Stylegan3: Curated Example Images},''
  \url{https://nvlabs-fi-cdn.nvidia.com/stylegan3/images/}, 2021.

\bibitem{karras2019stylebased}
T.~Karras, S.~Laine, and T.~Aila, ``{A Style-Based Generator Architecture for
  Generative Adversarial Networks},'' \emph{Proceedings of the IEEE/CVF
  Conference on Computer Vision and Pattern Recognition}, pp. 4401--4410, June
  2019, {Long Beach, CA}.

\bibitem{karras2020analyzing}
T.~Karras, S.~Laine, M.~Aittala, J.~Hellsten, J.~Lehtinen, and T.~Aila,
  ``{Analyzing and Improving the Image Quality of StyleGAN},''
  \emph{Proceedings of the IEEE/CVF Conference on Computer Vision and Pattern
  Recognition}, pp. 8110--8119, June 2020, {Virtual}.

\bibitem{nagano2019deep}
K.~Nagano, H.~Luo, Z.~Wang, J.~Seo, J.~Xing, L.~Hu, L.~Wei, and H.~Li, ``{Deep
  Face Normalization},'' \emph{ACM Transactions on Graphics}, vol.~38, no.~6,
  pp. 1--16, November 2019, {New York, NY}.

\bibitem{brock2019large}
A.~Brock, J.~Donahue, and K.~Simonyan, ``{Large Scale GAN Training for High
  Fidelity Natural Image Synthesis},'' \emph{Proceedings of the International
  Conference on Learning Representations}, May 2019, {New Orleans, LA}.

\bibitem{karras2020training}
T.~Karras, M.~Aittala, J.~Hellsten, S.~Laine, J.~Lehtinen, and T.~Aila,
  ``{Training Generative Adversarial Networks with Limited Data},''
  \emph{Advances in Neural Information Processing Systems}, vol.~33, pp.
  12\,104--12\,114, December 2020, {Virtual}.

\bibitem{karras2021aliasfree}
T.~Karras, M.~Aittala, S.~Laine, E.~H{\"a}rk{\"o}nen, J.~Hellsten, J.~Lehtinen,
  and T.~Aila, ``{Alias-Free Generative Adversarial Networks},'' \emph{Advances
  in Neural Information Processing Systems}, December 2021, {Virtual}.

\bibitem{vahdat2021scorebased}
A.~Vahdat, K.~Kreis, and J.~Kautz, ``{Score-based Generative Modeling in Latent
  Space},'' \emph{Advances in Neural Information Processing Systems}, vol.~34,
  pp. 11\,287--11\,302, December 2021, {Virtual}.

\bibitem{song2021scorebased}
Y.~Song, J.~Sohl-Dickstein, D.~P. Kingma, A.~Kumar, S.~Ermon, and B.~Poole,
  ``{Score-Based Generative Modeling through Stochastic Differential
  Equations},'' \emph{Proceedings of the International Conference on Learning
  Representations}, May 2021, {Vienna, Austria}.

\bibitem{chan2021efficient}
E.~R. Chan, C.~Z. Lin, M.~A. Chan, K.~Nagano, B.~Pan, S.~D. Mello, O.~Gallo,
  L.~Guibas, J.~Tremblay, S.~Khamis, T.~Karras, and G.~Wetzstein, ``{Efficient
  Geometry-Aware 3D Generative Adversarial Networks},'' \emph{arXiv preprint
  arXiv:2112.07945}, December 2021.

\bibitem{esser2021taming}
P.~Esser, R.~Rombach, and B.~Ommer, ``{Taming Transformers for High-Resolution
  Image Synthesis},'' \emph{Proceedings of the IEEE/CVF Conference on Computer
  Vision and Pattern Recognition}, pp. 12\,873--12\,883, June 2021, {Virtual}.

\bibitem{ramesh2021zero}
A.~Ramesh, M.~Pavlov, G.~Goh, S.~Gray, C.~Voss, A.~Radford, M.~Chen, and
  I.~Sutskever, ``Zero-shot text-to-image generation,'' \emph{Proceedings of
  the International Conference on Machine Learning}, vol. 139, pp. 8821--8831,
  July 2021.

\bibitem{ramesh2022hierarchichal}
A.~Ramesh, P.~Dhariwal, A.~Nichol, C.~Chu, and M.~Chen, ``{Hierarchical
  Text-Conditional Image Generation with CLIP Latents},'' \emph{arXiv preprint
  arXiv:2204.06125}, April 2022.

\bibitem{gpt3}
T.~B. Brown, B.~Mann, N.~Ryder, M.~Subbiah, J.~Kaplan, P.~Dhariwal,
  A.~Neelakantan, P.~Shyam, G.~Sastry, A.~Askell, S.~Agarwal, A.~Herbert-Voss,
  G.~Krueger, T.~Henighan, R.~Child, A.~Ramesh, D.~M. Ziegler, J.~Wu,
  C.~Winter, C.~Hesse, M.~Chen, E.~Sigler, M.~Litwin, S.~Gray, B.~Chess,
  J.~Clark, C.~Berner, S.~McCandlish, A.~Radford, I.~Sutskever, and D.~Amodei,
  ``{Language Models are Few-Shot Learners},'' \emph{Advances in Neural
  Information Processing Systems}, pp. 1--25, December 2020, {Virtual}.

\bibitem{wu2017deep}
Y.~Wu, W.~Abd-Almageed, and P.~Natarajan, ``{Deep Matching and Validation
  Network: An End-to-End Solution to Constrained Image Splicing Localization
  and Detection},'' \emph{Proceedings of the ACM International Conference on
  Multimedia}, pp. 1480--1502, October 2017, {Mountain View, CA}.

\bibitem{he2016deep}
K.~He, X.~Zhang, S.~Ren, and J.~Sun, ``{Deep Residual Learning for Image
  Recognition},'' \emph{Proceedings of the IEEE Conference on Computer Vision
  and Pattern Recognition}, pp. 770--778, June 2016, {Las Vegas, NV}.

\bibitem{liu2019adversarial}
Y.~Liu, X.~Zhu, X.~Zhao, and Y.~Cao, ``{Adversarial Learning for Constrained
  Image Splicing Detection and Localization Based on Atrous Convolution},''
  \emph{IEEE Transactions on Information Forensics and Security}, vol.~14,
  no.~10, pp. 2551--2566, March 2019.

\bibitem{bartusiak2019splicing}
E.~R. {Bartusiak}, S.~K. {Yarlagadda}, D.~{G\"{u}era}, P.~{Bestagini},
  S.~{Tubaro}, F.~M. {Zhu}, and E.~J. {Delp}, ``{Splicing Detection and
  Localization In Satellite Imagery Using Conditional GANs},''
  \emph{Proceedings of the IEEE Conference on Multimedia Information Processing
  and Retrieval}, pp. 91--96, March 2019, {San Jose, CA}.

\bibitem{bartusiak2019adversarial}
E.~R. Bartusiak, ``{An Adversarial Approach to Spliced Forgery Detection and
  Localization in Satellite Imagery},'' Master's thesis, Purdue University,
  West Lafayette, IN, USA, May 2019.

\bibitem{yarlagadda2019shadow}
S.~K. {Yarlagadda}, D.~{G\"{u}era}, D.~M. {Montserrat}, F.~M. {Zhu}, E.~J.
  {Delp}, P.~{Bestagini}, and S.~{Tubaro}, ``{Shadow Removal Detection and
  Localization for Forensics Analysis},'' \emph{Proceedings of the IEEE
  International Conference on Acoustics, Speech and Signal Processing}, pp.
  2677--2681, May 2019, {Brighton, UK}.

\bibitem{yarlagadda2018satellite}
S.~K. {Yarlagadda}, D.~{G\"{u}era}, P.~{Bestagini}, F.~M. {Zhu}, S.~{Tubaro},
  and E.~J. Delp, ``{Satellite Image Forgery Detection and Localization Using
  GAN and One-Class Classifier},'' \emph{Proceedings of the IS\&T International
  Symposium on Electronic Imaging: Media Watermarking, Security, and
  Forensics}, pp. 214--1--214--9, January 2018, {Burlingame, CA}.

\bibitem{kumar2019image}
M.~Kumar and S.~Srivastava, ``{Image Authentication by Assessing Manipulations
  Using Illumination},'' \emph{Multimedia Tools and Applications}, vol.~78,
  no.~9, pp. 12\,451--12\,463, May 2019.

\bibitem{zhu2021face}
X.~Zhu, H.~Wang, H.~Fei, Z.~Lei, and S.~Z. Li, ``{Face Forgery Detection by 3D
  Decomposition},'' \emph{Proceedings of the IEEE/CVF Conference on Computer
  Vision and Pattern Recognition}, pp. 2929--2939, June 2021, {Virtual}.

\bibitem{niu2021image}
Y.~Niu, B.~Tondi, Y.~Zhao, R.~Ni, and M.~Barni, ``{Image Splicing Detection,
  Localization and Attribution via JPEG Primary Quantization Matrix Estimation
  and Clustering},'' \emph{IEEE Transactions on Information Forensics and
  Security}, vol.~16, pp. 5397--5412, November 2021.

\bibitem{kwon2021catnet}
M.-J. Kwon, I.-J. Yu, S.-H. Nam, and H.-K. Lee, ``{CAT-Net: Compression
  Artifact Tracing Network for Detection and Localization of Image Splicing},''
  \emph{Proceedings of the IEEE Winter Conference on Applications of Computer
  Vision}, pp. 375--384, January 2021, {Virtual}.

\bibitem{bonettini2019image}
N.~Bonettini, D.~G\"{u}era, L.~Bondi, P.~Bestagini, E.~J. Delp, and S.~Tubaro,
  ``{Image Anonymization Detection with Deep Handcrafted Features},''
  \emph{Proceedings of the IEEE International Conference on Image Processing},
  September 2019, {Taipei, Taiwan}.

\bibitem{guera2018reliability}
D.~G\"{u}era, S.~K. Yarlagadda, P.~Bestagini, F.~Zhu, S.~Tubaro, and E.~J.
  Delp, ``{Reliability Map Estimation For CNN-Based Camera Model
  Attribution},'' \emph{Proceedings of the IEEE Winter Conference on
  Applications of Computer Vision}, pp. 964--973, March 2018, {Lake Tahoe, NV}.

\bibitem{charitidis2021operationwise}
P.~Charitidis, G.~Kordopatis-Zilos, S.~Papadopoulos, and I.~Kompatsiaris,
  ``{Operation-wise Attention Network for Tampering Localization Fusion},''
  \emph{Proceedings of the International Conference on Content-Based Multimedia
  Indexing}, pp. 1--6, June 2021, {Lille, France}.

\bibitem{barni2021copy}
M.~Barni, Q.-T. Phan, and B.~Tondi, ``{Copy Move Source-Target Disambiguation
  Through Multi-Branch CNNs},'' \emph{IEEE Transactions on Information
  Forensics and Security}, vol.~16, pp. 1825--1840, January 2021.

\bibitem{cozzolino2019noiseprint}
D.~Cozzolino and L.~Verdoliva, ``{Noiseprint: A CNN-Based Camera Model
  Fingerprint},'' \emph{IEEE Transactions on Information Forensics and
  Security}, vol.~15, pp. 144--159, May 2019.

\bibitem{wu2019mantranet}
Y.~Wu, W.~AbdAlmageed, and P.~Natarajan, ``{ManTra-Net: Manipulation Tracing
  Network for Detection and Localization of Image Forgeries With Anomalous
  Features},'' \emph{Proceedings of the IEEE/CVF Conference on Computer Vision
  and Pattern Recognition}, pp. 9543--9552, June 2019, {Long Beach, CA}.

\bibitem{chen2021image}
X.~Chen, C.~Dong, J.~Ji, J.~Cao, and X.~Li, ``{Image Manipulation Detection by
  Multi-View Multi-Scale Supervision},'' \emph{Proceedings of the IEEE/CVF
  International Conference on Computer Vision}, pp. 14\,185--14\,193, October
  2021, {Virtual}.

\bibitem{cannas2021open}
E.~Cannas, S.~Baireddy, E.~R. Bartusiak, S.~K. Yarlagadda, D.~M. Montserrat,
  P.~Bestagini, S.~Tubaro, and E.~J. Delp, ``{Open-Set Source Attribution for
  Panchromatic Satellite Imagery},'' \emph{Proceedings of the IEEE
  International Conference on Image Processing}, pp. 3038--3042, September
  2021, {Virtual}.

\bibitem{horvath2021manipulation}
J.~Horv\'{a}th, S.~Baireddy, H.~Hao, D.~M. Montserrat, and E.~J. Delp,
  ``{Manipulation Detection in Satellite Images Using Vision Transformer},''
  \emph{Proceedings of the IEEE/CVF Conference on Computer Vision and Pattern
  Recognition Workshops}, pp. 1032--1041, June 2021, {Virtual}.

\bibitem{horvath2020manipulation}
J.~Horv\'{a}th, D.~M. Montserrat, H.~Hao, and E.~J. Delp, ``{Manipulation
  Detection in Satellite Images Using Deep Belief Networks},''
  \emph{Proceedings of the IEEE/CVF Conference on Computer Vision and Pattern
  Recognition Workshops}, pp. 2832--2840, June 2020, {Virtual}.

\bibitem{horvath2019manipulation}
J.~Horv\'{a}th, D.~G\"{u}era, S.~K. Yarlagadda, P.~Bestagini, F.~M. Zhu,
  S.~Tubaro, and E.~J. Delp, ``{Anomaly-Based Manipulation Detection in
  Satellite Images},'' \emph{Proceedings of the IEEE/CVF Conference on Computer
  Vision and Pattern Recognition Workshops}, pp. 62--71, June 2019, {Long
  Beach, CA}.

\bibitem{montserrat2020generative}
D.~M. Montserrat, J.~Horv\'{a}th, S.~K. Yarlagadda, F.~Zhu, and E.~J. Delp,
  ``{Generative Autoregressive Ensembles for Satellite Imagery Manipulation
  Detection},'' \emph{Proceedings of the IEEE International Workshop on
  Information Forensics and Security}, pp. 1--6, December 2020, {Virtual}.

\bibitem{giudice2021fighting}
O.~Giudice, L.~Guarnera, and S.~Battiato, ``{Fighting Deepfakes by Detecting
  GAN DCT Anomalies},'' \emph{Journal of Imaging}, vol.~7, no.~8, July 2021.

\bibitem{wang2020cnngenerated}
S.-Y. Wang, O.~Wang, R.~Zhang, A.~Owens, and A.~A. Efros, ``{CNN-Generated
  Images Are Surprisingly Easy to Spot... for Now},'' \emph{Proceedings of the
  IEEE/CVF Conference on Computer Vision and Pattern Recognition}, pp.
  8695--8704, June 2020, {Virtual}.

\bibitem{girish2021towards}
S.~Girish, S.~Suri, S.~S. Rambhatla, and A.~Shrivastava, ``{Towards Discovery
  and Attribution of Open-World GAN Generated Images},'' \emph{Proceedings of
  the IEEE/CVF International Conference on Computer Vision}, pp.
  14\,094--14\,103, October 2021, {Virtual}.

\bibitem{cozzolino2021towards}
D.~Cozzolino, D.~Gragnaniello, G.~Poggi, and L.~Verdoliva, ``{Towards Universal
  GAN Image Detection},'' \emph{Proceedings of the International Conference on
  Visual Communications and Image Processing}, pp. 1--5, December 2021,
  {Munich, Germany}.

\bibitem{guarnera2020fighting}
L.~Guarnera, O.~Giudice, and S.~Battiato, ``{Fighting Deepfake by Exposing the
  Convolutional Traces on Images},'' \emph{IEEE Access}, vol.~8, pp.
  165\,085--165\,098, January 2020.

\bibitem{hu2021exposing}
S.~Hu, Y.~Li, and S.~Lyu, ``{Exposing GAN-Generated Faces Using Inconsistent
  Corneal Specular Highlights},'' \emph{Proceedings of the IEEE International
  Conference on Acoustics, Speech and Signal Processing}, pp. 2500--2504, June
  2021, {Toronto, Canada}.

\bibitem{guo2021eyes}
H.~Guo, S.~Hu, X.~Wang, M.-C. Chang, and S.~Lyu, ``{Eyes Tell All: Irregular
  Pupil Shapes Reveal GAN-generated Faces},'' \emph{arXiv preprint
  arXiv:2109.00162}, October 2021.

\bibitem{kaur2020image}
H.~Kaur and N.~Jindal, ``{Image and Video Forensics: A Critical Survey},''
  \emph{Wireless Personal Communications}, vol. 112, pp. 1281--1302, January
  2020.

\bibitem{guera2018deepfake}
D.~G\"{u}era and E.~Delp, ``{Deepfake Video Detection Using Recurrent Neural
  Networks},'' \emph{Proceedings of the IEEE International Conference on
  Advanced Video and Signal-based Surveillance}, pp. 1--6, November 2018,
  {Auckland, New Zealand}.

\bibitem{guera2019media}
D.~G\"{u}era, ``{Media Forensics Using Machine Learning Approaches},'' Ph.D.
  dissertation, Purdue University, West Lafayette, IN, USA, October 2019.

\bibitem{ravi2014compression}
H.~Ravi, A.~V. Subramanyam, G.~Gupta, and B.~A. Kumar, ``{Compression Noise
  Based Video Forgery Detection},'' \emph{Proceedings of the IEEE International
  Conference on Image Processing}, pp. 5352--5356, October 2014, {Paris,
  France}.

\bibitem{mahfoudi2022statistical}
G.~Mahfoudi, F.~Retraint, F.~Morain-Nicolier, and M.~M. Pic, ``{Statistical
  H.264 Double Compression Detection Method Based on DCT Coefficients},''
  \emph{IEEE Access}, vol.~10, pp. 4271--4283, January 2022.

\bibitem{sullivan2012overview}
G.~J. Sullivan, J.-R. Ohm, W.-J. Han, and T.~Wiegand, ``{Overview of the High
  Efficiency Video Coding (HEVC) Standard},'' \emph{IEEE Transactions on
  Circuits and Systems for Video Technology}, vol.~22, no.~12, pp. 1649--1668,
  September 2012.

\bibitem{uddin2022double}
K.~Uddin, Y.~Yang, and B.~T. Oh, ``{Double Compression Detection in HEVC-coded
  Video with the Same Coding Parameters Using Picture Partitioning
  Information},'' \emph{Signal Processing: Image Communication}, vol. 103, p.
  116638, April 2022.

\bibitem{kang2021edge}
G.-Y. Kang, Y.-P. Feng, R.-K. Wang, and Z.-M. Lu, ``{Edge and Feature Points
  Based Video Intra-frame Passive-blind Copy-paste Forgery Detection},''
  \emph{Journal of Network Intelligence}, vol.~6, no.~3, pp. 637--645, August
  2021.

\bibitem{lowe1999object}
D.~Lowe, ``{Object Recognition from Local Scale-Invariant Features},''
  \emph{Proceedings of the IEEE International Conference on Computer Vision},
  vol.~2, pp. 1150--1157, September 1999, {Kerkyra, Greece}.

\bibitem{singh2022chroma}
G.~Singh and K.~Singh, ``{Chroma Key Foreground Forgery Detection Under Various
  Attacks in Digital Video Based on Frame Edge Identification},''
  \emph{Multimedia Tools and Applications}, vol.~81, no.~1, pp. 1419--1446,
  January 2022.

\bibitem{canny1986computational}
J.~Canny, ``{A Computational Approach to Edge Detection},'' \emph{IEEE
  Transactions on Pattern Analysis and Machine Intelligence}, vol.~8, no.~6,
  pp. 679--698, November 1986.

\bibitem{adoui2021video}
A.~Adoui El~Ouadrhiri, S.~Jai-Andaloussi, and O.~Ouchetto, ``{Video Block and
  FABEMD Features for an Effective and Fast Method of Reporting Near-Duplicate
  and Mirroring Videos},'' \emph{Journal of Big Data}, vol.~8, no.~1, p. 138,
  October 2021.

\bibitem{rumelhart1986learning}
D.~E. Rumelhart, G.~E. Hinton, and R.~J. Williams, ``{Learning Representations
  by Back-Propagating Errors},'' \emph{Nature}, vol. 323, no. 6088, pp.
  533--536, October 1986.

\bibitem{cozzolino2021idreveal}
D.~Cozzolino, A.~R\"ossler, J.~Thies, M.~Nie{\ss}ner, and L.~Verdoliva,
  ``{ID-Reveal: Identity-Aware DeepFake Video Detection},'' \emph{Proceedings
  of the IEEE/CVF International Conference on Computer Vision}, pp.
  15\,108--15\,117, October 2021, {Virtual}.

\bibitem{guo2020towards}
J.~Guo, X.~Zhu, Y.~Yang, F.~Yang, Z.~Lei, and S.~Z. Li, ``{Towards Fast,
  Accurate and Stable 3D Dense Face Alignment},'' \emph{Proceedings of the
  European Conference on Computer Vision}, pp. 152--168, August 2020,
  {Virtual}.

\bibitem{bonettini2021video}
N.~Bonettini, E.~D. Cannas, S.~Mandelli, L.~Bondi, P.~Bestagini, and S.~Tubaro,
  ``{Video Face Manipulation Detection Through Ensemble of CNNs},''
  \emph{Proceedings of the IEEE International Conference on Pattern
  Recognition}, pp. 5012--5019, January 2021, {Milan, Italy}.

\bibitem{tan2019efficientnet}
M.~Tan and Q.~Le, ``{EfficientNet: Rethinking Model Scaling for Convolutional
  Neural Networks},'' \emph{Proceedings of the International Conference on
  Machine Learning}, vol.~97, pp. 6105--6114, June 2019, {Long Beach, CA}.

\bibitem{montserrat2020deepfakes}
D.~M. Montserrat, H.~Hao, S.~K. Yarlagadda, S.~Baireddy, R.~Shao,
  J.~Horv\'{a}th, E.~Bartusiak, J.~Yang, D.~G\"{u}era, F.~Zhu, and E.~J. Delp,
  ``{Deepfakes Detection with Automatic Face Weighting},'' \emph{Proceedings of
  the IEEE/CVF Conference on Computer Vision and Pattern Recognition
  Workshops}, pp. 2851--2859, June 2020, {Virtual}.

\bibitem{hao2022deepfake}
H.~Hao, E.~R. Bartusiak, D.~G\"{u}era, D.~Mas, S.~Baireddy, Z.~Xiang, S.~K.
  Yarlagadda, R.~Shao, J.~Horv\'{a}th, J.~Yang, F.~Zhu, and E.~J. Delp,
  ``{Deepfake Detection Using Multiple Data Modalities},'' in \emph{{Handbook
  of Digital Face Manipulation and Detection - From DeepFakes to Morphing
  Attacks, Series on Advances in Computer Vision and Pattern
  Recognition}}.\hskip 1em plus 0.5em minus 0.4em\relax Springer, March 2022,
  vol.~1, pp. 235--254.

\bibitem{zhang2016joint}
K.~Zhang, Z.~Zhang, Z.~Li, and Y.~Qiao, ``{Joint Face Detection and Alignment
  Using Multitask Cascaded Convolutional Networks},'' \emph{IEEE Signal
  Processing Letters}, vol.~23, no.~10, pp. 1499--1503, August 2016.

\bibitem{vaswani2017attention}
A.~Vaswani, N.~Shazeer, N.~Parmar, J.~Uszkoreit, L.~Jones, A.~N. Gomez,
  L.~Kaiser, and I.~Polosukhin, ``{Attention is All You Need},'' \emph{Advances
  in Neural Information Processing Systems}, vol.~30, pp. 6000--6010, December
  2017, {Long Beach, CA}.

\bibitem{guhagarkar2021novel}
N.~Guhagarkar, S.~Desai, S.~Vaishampayan, and A.~Save, ``{A Novel Approach to
  Detect Low Quality Deepfake Videos},'' \emph{Sentimental Analysis and Deep
  Learning}, vol. 1408, no.~1, pp. 879--891, October 2021.

\bibitem{hochreiter1997long}
S.~Hochreiter and J.~Schmidhuber, ``{Long Short-Term Memory},'' \emph{Neural
  Computation}, vol.~9, no.~8, p. 1735–1780, November 1997.

\bibitem{marcon2021detection}
F.~Marcon, C.~Pasquini, and G.~Boato, ``{Detection of Manipulated Face Videos
  over Social Networks: A Large-Scale Study},'' \emph{Journal of Imaging},
  vol.~7, p. 193, September 2021.

\bibitem{bondi2020training}
L.~Bondi, E.~Daniele~Cannas, P.~Bestagini, and S.~Tubaro, ``{Training
  Strategies and Data Augmentations in CNN-Based DeepFake Video Detection},''
  \emph{Proceedings of the IEEE International Workshop on Information Forensics
  and Security}, pp. 1--6, December 2020, {Shanghai, China}.

\bibitem{hinton2011transforming}
G.~E. Hinton, A.~Krizhevsky, and S.~D. Wang, ``{Transforming Auto-Encoders},''
  in \emph{{Artificial Neural Networks and Machine Learning}}.\hskip 1em plus
  0.5em minus 0.4em\relax Springer Berlin Heidelberg, 2011, pp. 44--51.

\bibitem{nguyen2022capsuleforensics}
H.~H. Nguyen, J.~Yamagishi, and I.~Echizen, ``{Capsule-Forensics Networks for
  Deepfake Detection},'' in \emph{{Handbook of Digital Face Manipulation and
  Detection - From DeepFakes to Morphing Attacks, Series on Advances in
  Computer Vision and Pattern Recognition}}.\hskip 1em plus 0.5em minus
  0.4em\relax Springer, March 2022, vol.~1, pp. 275--301.

\bibitem{mazzia2021efficientcapsnet}
V.~Mazzia, F.~Salvetti, and M.~Chiaberge, ``{Efficient-CapsNet: Capsule Network
  with Self-Attention Routing},'' \emph{Scientific Reports, Springer Nature},
  vol.~11, p. 14634, July 2021.

\bibitem{huang2020dacapsnet}
W.~Huang and F.~Zhou, ``{DA-CapsNet: Dual Attention Mechanism Capsule
  Network},'' \emph{Scientific Reports, Springer Nature}, vol.~10, p. 11383,
  July 2020.

\bibitem{alamayreh2021detection}
O.~Alamayreh and M.~Barni, ``{Detection of GAN-Synthesized Street Videos},''
  \emph{Proceedings of the European Signal Processing Conference}, pp.
  811--815, September 2021, {Virtual}.

\bibitem{bevinamarad2020audio}
P.~R. Bevinamarad and M.~Shirldonkar, ``{Audio Forgery Detection Techniques:
  Present and Past Review},'' \emph{Proceedings of the International Conference
  on Trends in Electronics and Informatics}, pp. 613--618, June 2020,
  {Tirunelveli, India}.

\bibitem{buchholz2009microphone}
R.~Buchholz, C.~Kraetzer, and J.~Dittmann, ``{Microphone Classification Using
  Fourier Coefficients},'' \emph{International Workshop on Information Hiding},
  pp. 235--246, June 2009.

\bibitem{ikram2010digital}
S.~Ikram and H.~Malik, ``{Digital Audio Forensics Using Background Noise},''
  \emph{Proceedings of the IEEE International Conference on Multimedia and
  Expo}, pp. 106--110, July 2010, {Singapore}.

\bibitem{cuccovillo2013audio}
L.~Cuccovillo, S.~Mann, M.~Tagliasacchi, and P.~Aichroth, ``{Audio Tampering
  Detection via Microphone Classification},'' \emph{Proceedings of the IEEE
  International Workshop on Multimedia Signal Processing}, pp. 177--182,
  October 2013, {Pula, Italy}.

\bibitem{gupta2011current}
S.~Gupta, S.~Cho, and C.-C.~J. Kuo, ``{Current Developments and Future Trends
  in Audio Authentication},'' \emph{IEEE MultiMedia}, vol.~19, no.~1, pp.
  50--59, December 2011.

\bibitem{zhao2014audio}
H.~Zhao, Y.~Chen, R.~Wang, and H.~Malik, ``{Audio Source Authentication and
  Splicing Detection Using Acoustic Environmental Signature},''
  \emph{Proceedings of the ACM Workshop on Information Hiding and Multimedia
  Security}, pp. 159--164, June 2014, {Salzburg, Austria}.

\bibitem{yang2010detecting}
R.~Yang, Y.~Q. Shi, and J.~Huang, ``{Detecting Double Compression of Audio
  Signal},'' in \emph{{Media Forensics and Security II}}.\hskip 1em plus 0.5em
  minus 0.4em\relax SPIE, January 2010, vol. 7541, pp. 200--209.

\bibitem{liu2010detection}
Q.~Liu, A.~H. Sung, and M.~Qiao, ``{Detection of Double MP3 Compression},''
  \emph{Cognitive Computation}, vol.~2, no.~4, pp. 291--296, May 2010.

\bibitem{luo2016detection}
D.~Luo, R.~Yang, B.~Li, and J.~Huang, ``{Detection of Double Compressed AMR
  Audio Using Stacked Autoencoder},'' \emph{IEEE Transactions on Information
  Forensics and Security}, vol.~12, no.~2, pp. 432--444, October 2016.

\bibitem{iso-mp3}
``{ISO/IEC 13818-7:1997 - Information Technology -- Generic Coding of Moving
  Pictures and Associated Audio Information -- Part 7: Advanced Audio Coding
  (AAC)},'' \url{https://www.iso.org/standard/25040.html}, 1997.

\bibitem{brandenburg1999mp3}
K.~Brandenburg, ``{MP3 and AAC Explained},'' \emph{Proceedings of the AES
  International Conference on High-Quality Audio Coding}, September 1999,
  {Signa, Italy}.

\bibitem{xiang2022forensic}
Z.~Xiang, P.~Bestagini, S.~Tubaro, and E.~J. Delp, ``{Forensic Analysis and
  Localization of Multiply Compressed MP3 Audio Using Transformers},''
  \emph{arXiv preprint arXiv:2203.16499}, March 2022.

\bibitem{yan2019robust}
Q.~Yan, R.~Yang, and J.~Huang, ``{Robust Copy-Move Detection of Speech
  Recording Using Similarities of Pitch and Formant},'' \emph{IEEE Transactions
  on Information Forensics and Security}, vol.~14, no.~9, pp. 2331--2341,
  January 2019.

\bibitem{akdeniz2021detection}
F.~Akdeniz and Y.~Becerikli, ``{Detection of Copy-Move Forgery in Audio Signal
  with Mel Frequency and Delta-Mel Frequency Kepstrum Coefficients},''
  \emph{Proceedings of the Innovations in Intelligent Systems and Applications
  Conference}, pp. 1--6, October 2021, {Elazig, Turkey}.

\bibitem{kim2021conditional}
J.~Kim, J.~Kong, and J.~Son, ``{Conditional Variational Autoencoder with
  Adversarial Learning for End-to-End Text-to-Speech},'' \emph{Proceedings of
  the International Conference on Machine Learning}, vol. 139, pp. 5530--5540,
  July 2021, {Virtual}.

\bibitem{wang2021prosody}
T.~Wang, R.~Fu, J.~Yi, J.~Tao, Z.~Wen, C.~Qiang, and S.~Wang, ``{Prosody and
  Voice Factorization for Few-Shot Speaker Adaptation in the Challenge M2voc
  2021},'' \emph{Proceedings of the IEEE International Conference on Acoustics,
  Speech and Signal Processing}, pp. 8603--8607, June 2021, {Toronto, Canada}.

\bibitem{popov2021gradtts}
V.~Popov, I.~Vovk, V.~Gogoryan, T.~Sadekova, and M.~Kudinov, ``{Grad-TTS: A
  Diffusion Probabilistic Model for Text-to-Speech},'' \emph{Proceedings of the
  International Conference on Machine Learning}, vol. 139, pp. 8599--8608, July
  2021, {Virtual}.

\bibitem{zhou2021seen}
K.~Zhou, B.~Sisman, R.~Liu, and H.~Li, ``{Seen and Unseen Emotional Style
  Transfer for Voice Conversion with A New Emotional Speech Dataset},''
  \emph{Proceedings of the IEEE International Conference on Acoustics, Speech
  and Signal Processing}, pp. 920--924, June 2021, {Toronto, Canada}.

\bibitem{yamagishi2021asvspoof}
J.~Yamagishi, X.~Wang, M.~Todisco, M.~Sahidullah, J.~Patino, A.~Nautsch,
  X.~Liu, K.~A. Lee, T.~Kinnunen, N.~Evans, and H.~Delgado, ``{ASVspoof 2021:
  Accelerating Progress in Spoofed and Deepfake Speech Detection},''
  \emph{Proceedings of the Automatic Speaker Verification and Spoofing
  Countermeasures Challenge}, pp. 47--54, September 2021.

\bibitem{yi2022add}
J.~Yi, R.~Fu, J.~Tao, S.~Nie, H.~Ma, C.~Wang, T.~Wang, Z.~Tian, Y.~Bai, C.~Fan
  \emph{et~al.}, ``{Add 2022: The First Audio Deep Synthesis Detection
  Challenge},'' \emph{arXiv preprint arXiv:2202.08433}, February 2022.

\bibitem{todisco2017constant}
M.~Todisco, H.~Delgado, and N.~Evans, ``{Constant Q Cepstral Coefficients: A
  Spoofing Countermeasure for Automatic Speaker Verification},'' \emph{Computer
  Speech \& Language}, vol.~45, pp. 516--535, September 2017.

\bibitem{hassan2021voice}
F.~Hassan and A.~Javed, ``{Voice Spoofing Countermeasure for Synthetic Speech
  Detection},'' \emph{Proceedings of the International Conference on Artificial
  Intelligence}, pp. 209--212, April 2021, {Settat, Morocco}.

\bibitem{subramani2020learning}
N.~Subramani and D.~Rao, ``{Learning Efficient Representations for Fake Speech
  Detection},'' \emph{Proceedings of the AAAI Conference on Artificial
  Intelligence}, vol.~34, no.~04, pp. 5859--5866, April 2020, {New York, NY}.

\bibitem{yang2020longterm}
J.~Yang and R.~K. Das, ``{Long-term High Frequency Features for Synthetic
  Speech Detection},'' \emph{Digital Signal Processing}, vol.~97, p. 102622,
  February 2020.

\bibitem{das2019long}
R.~K. Das, J.~Yang, and H.~Li, ``{Long Range Acoustic Features for Spoofed
  Speech Detection},'' \emph{Proceedings of Interspeech}, pp. 1058--1062,
  September 2019, {Graz, Austria}.

\bibitem{li2022longterm}
J.~Li, H.~Wang, P.~He, S.~M. Abdullahi, and B.~Li, ``{Long-term Variable Q
  transform: A Novel Time-frequency Transform Algorithm for Synthetic Speech
  Detection},'' \emph{Digital Signal Processing}, vol. 120, p. 103256, January
  2022.

\bibitem{li2021replay}
X.~Li, N.~Li, C.~Weng, X.~Liu, D.~Su, D.~Yu, and H.~Meng, ``{Replay and
  Synthetic Speech Detection with Res2Net Architecture},'' \emph{Proceedings of
  the IEEE International Conference on Acoustics, Speech and Signal
  Processing}, pp. 6354--6358, June 2021, {Toronto, Canada}.

\bibitem{mittal2021automatic}
A.~Mittal and M.~Dua, ``{Automatic Speaker Verification Systems and Spoof
  Detection Techniques: Review and Analysis},'' \emph{International Journal of
  Speech Technology}, p. 105–134, August 2021.

\bibitem{choi2021comparison}
K.~Choi, G.~Fazekas, K.~Cho, and M.~Sandler, ``{A Comparison of Audio Signal
  Preprocessing Methods for Deep Neural Networks on Music Tagging},''
  \emph{arXiv preprint arXiv:1709.01922}, February 2021.

\bibitem{mel}
S.~Stevens, J.~Volkmann, and E.~Newman, ``{A Scale for the Measurement of the
  Psychological Magnitude Pitch},'' \emph{Journal of the Acoustical Society of
  America}, vol.~8, pp. 185--190, June 1937.

\bibitem{bartusiak2021frequency}
E.~R. Bartusiak and E.~J. Delp, ``{Frequency Domain-Based Detection of
  Generated Audio},'' \emph{Proceedings of the IS\&T International Symposium on
  Electronic Imaging: Media Watermarking, Security, and Forensics}, pp. 273--1
  -- 273--7, January 2021, {Burlingame, CA}.

\bibitem{bartusiak2021synthesized}
E.~R. Bartusiak and E.~J. Delp, ``{Synthesized Speech Detection Using
  Convolutional Transformer-Based Spectrogram Analysis},'' \emph{Proceedings of
  the IEEE Asilomar Conference on Signals, Systems, and Computers}, October
  2021, {Asilomar, CA}.

\bibitem{khochare2021deep}
J.~Khochare, C.~Joshi, B.~Yenarkar, S.~Suratkar, and F.~Kazi, ``{A Deep
  Learning Framework for Audio Deepfake Detection},'' \emph{Arabian Journal for
  Science and Engineering}, vol.~47, p. 3447–3458, November 2021.

\bibitem{hua2021towards}
G.~Hua, A.~B.~J. Teoh, and H.~Zhang, ``{Towards End-to-End Synthetic Speech
  Detection},'' \emph{IEEE Signal Processing Letters}, vol.~28, pp. 1265--1269,
  June 2021.

\bibitem{szegedy2015going}
C.~Szegedy, W.~Liu, Y.~Jia, P.~Sermanet, S.~Reed, D.~Anguelov, D.~Erhan,
  V.~Vanhoucke, and A.~Rabinovich, ``{Going Deeper with Convolutions},''
  \emph{Proceedings of the IEEE/CVF Conference on Computer Vision and Pattern
  Recognition}, pp. 1--9, June 2015, {Boston}.

\bibitem{chintha2020recurrent}
A.~Chintha, B.~Thai, S.~J. Sohrawardi, K.~Bhatt, A.~Hickerson, M.~Wright, and
  R.~Ptucha, ``{Recurrent Convolutional Structures for Audio Spoof and Video
  Deepfake Detection},'' \emph{IEEE Journal of Selected Topics in Signal
  Processing}, vol.~14, no.~5, pp. 1024--1037, June 2020.

\bibitem{elkasrawi2014printer}
S.~Elkasrawi and F.~Shafait, ``{Printer Identification Using Supervised
  Learning for Document Forgery Detection},'' \emph{Proceedings of the IAPR
  International Workshop on Document Analysis Systems}, pp. 146--150, April
  2014, {Tours, France}.

\bibitem{jawahar2020automatic}
G.~Jawahar, M.~Abdul-Mageed, and L.~V.~S. Lakshmanan, ``{Automatic Detection of
  Machine Generated Text: A Critical Survey},'' \emph{arXiv preprint
  arXiv:2011.01314}, November 2020.

\bibitem{aldwairi2018detecting}
M.~Aldwairi and A.~Alwahedi, ``{Detecting Fake News in Social Media
  Networks},'' \emph{Procedia Computer Science}, vol. 141, pp. 215--222,
  November 2018.

\bibitem{ahmad2020fake}
I.~Ahmad, M.~Yousaf, S.~Yousaf, and M.~O. Ahmad, ``{Fake News Detection Using
  Machine Learning Ensemble Methods},'' \emph{Complexity}, October 2020.

\bibitem{bakhtin2021residual}
A.~Bakhtin, Y.~Deng, S.~Gross, M.~Ott, M.~Ranzato, and A.~Szlam, ``{Residual
  Energy-Based Models for Text},'' \emph{Journal of Machine Learning Research},
  vol.~22, pp. 40--1, January 2021.

\bibitem{gpt2}
A.~Radford, J.~Wu, R.~Child, D.~Luan, D.~Amodei, and I.~Sutskever, ``{Language
  Models are Unsupervised Multitask Learners},'' \emph{Technical Report}, 2019,
  {OpenAI Inc., San Francisco, CA}.

\bibitem{bert2019}
J.~Devlin, M.-W. Chang, K.~Lee, and K.~Toutanova, ``{BERT: Pre-Training of Deep
  Bidirectional Transformers for Language Understanding},'' \emph{Proceedings
  of the Conference of the North American Chapter of the Association for
  Computational Linguistics: Human Language Technologies}, pp. 4171--4186, June
  2019, {Minneapolis, MN}.

\bibitem{zellers2019defending}
R.~Zellers, A.~Holtzman, H.~Rashkin, Y.~Bisk, A.~Farhadi, F.~Roesner, and
  Y.~Choi, ``{Defending Against Neural Fake News},'' \emph{Advances in Neural
  Information Processing Systems}, vol.~32, December 2019, {Vancouver, Canada}.

\bibitem{stamatatos2009survey}
E.~Stamatatos, ``{A Survey of Modern Authorship Attribution Methods},''
  \emph{Journal of the American Society for information Science and
  Technology}, vol.~60, no.~3, pp. 538--556, December 2009.

\bibitem{ding2017learning}
S.~H. Ding, B.~C. Fung, F.~Iqbal, and W.~K. Cheung, ``{Learning Stylometric
  Representations for Authorship Analysis},'' \emph{IEEE Transactions on
  Cybernetics}, vol.~49, no.~1, pp. 107--121, November 2017.

\bibitem{bhargava2013stylometric}
M.~Bhargava, P.~Mehndiratta, and K.~Asawa, ``{Stylometric Analysis for
  Authorship Attribution on Twitter},'' \emph{Proceedings of the International
  Conference on Big Data Analytics}, pp. 37--47, December 2013, {{Mysore,
  India}}.

\bibitem{kalgutkar2019code}
V.~Kalgutkar, R.~Kaur, H.~Gonzalez, N.~Stakhanova, and A.~Matyukhina, ``{Code
  Authorship Attribution: Methods and Challenges},'' \emph{Association for
  Computing Machinery Computing Surveys}, vol.~52, no.~1, pp. 1--36, February
  2019.

\bibitem{halvani2019assessing}
O.~Halvani, C.~Winter, and L.~Graner, ``{Assessing the Applicability of
  Authorship Verification Methods},'' \emph{Proceedings of the International
  Conference on Availability, Reliability and Security}, pp. 1--10, August
  2019, {Canterbury, United Kingdom}.

\bibitem{fabien2020bertaa}
M.~Fabien, E.~Villatoro-Tello, P.~Motlicek, and S.~Parida, ``{BertAA: BERT
  fine-tuning for Authorship Attribution},'' \emph{Proceedings of the
  International Conference on Natural Language Processing}, pp. 127--137,
  December 2020, {Patna, India}.

\bibitem{shrestha2017convolutional}
P.~Shrestha, S.~Sierra, F.~A. Gonz\'{a}lez, M.~Montes-y G\'{o}mez, P.~Rosso,
  and T.~Solorio, ``{Convolutional Neural Networks for Authorship Attribution
  of Short Texts},'' \emph{Proceedings of the Conference of the European
  Chapter of the Association for Computational Linguistics}, pp. 669--674,
  April 2017, {Valencia, Spain}.

\bibitem{gideon2018handwritten}
S.~J. Gideon, A.~Kandulna, A.~A. Kujur, A.~Diana, and K.~Raimond,
  ``{Handwritten Signature Forgery Detection using Convolutional Neural
  Networks},'' \emph{Procedia Computer Science}, vol. 143, pp. 978--987,
  November 2018.

\bibitem{beusekom2012text}
J.~van Beusekom, F.~Shafait, and T.~M. Breuel, ``{Text-Line Examination for
  Document Forgery Detection},'' \emph{International Journal on Document
  Analysis and Recognition}, vol.~16, p. 189–207, January 2012.

\bibitem{cruz2017local}
F.~Cruz, N.~Sidère, M.~Coustaty, V.~P. D'Andecy, and J.-M. Ogier, ``{Local
  Binary Patterns for Document Forgery Detection},'' \emph{Proceedings of the
  IAPR International Conference on Document Analysis and Recognition}, vol.~01,
  pp. 1223--1228, November 2017, {Kyoto, Japan}.

\bibitem{noble2006what}
W.~S. Noble, ``{What is a Support Vector Machine?}'' \emph{Nature
  Biotechnology}, vol.~24, December 2006.

\bibitem{rocha2017}
A.~Ferreira, L.~Bondi, L.~Baroffio, P.~Bestagini, J.~Huang, J.~A. dos Santos,
  S.~Tubaro, and A.~Rocha, ``{Data-Driven Feature Characterization Techniques
  for Laser Printer Attribution},'' \emph{IEEE Transactions on Information
  Forensics and Security}, vol.~12, no.~8, pp. 1860--1873, April 2017.

\bibitem{Bik2016prevalence}
E.~Bik, A.~Casadevall, and F.~Fang, ``{The Prevalence of Inappropriate Image
  Duplication},'' \emph{Biomedical Research Publications}, vol.~7, no.~3, June
  2016.

\bibitem{lonni2022correction}
L.~B. Onid, E.~Bik, J.~Heathers, and G.~Meyerowitz-Katz, ``{Correction of
  Scientific Literature: Too Little, Too Late!}'' \emph{PLOS Biology}, vol.~20,
  p. e3001572, March 2022.

\bibitem{daniel2022sila}
D.~Moreira, J.~a.~P. Cardenuto, R.~Shao, S.~Baireddy, D.~Cozzolino,
  D.~Gragnaniello, W.~Abd-Almageed, P.~Bestagini, S.~Tubaro, A.~Rocha,
  W.~Scheirer, L.~Verdoliva, and E.~J. Delp, ``{SILA: A System for Scientific
  Image Analysis},'' \emph{Under Preparation}.

\bibitem{zhuang2021graphical}
H.~Zhuang, T.-Y. Huang, and D.~E. Acuna, ``{Graphical Integrity Issues in Open
  Access Publications: Detection and Patterns of Proportional Ink
  Violations},'' \emph{PLOS Computational Biology}, vol.~17, no.~12, p.
  e1009650, December 2021.

\bibitem{vo2020where}
N.~Vo and K.~Lee, ``{Where Are the Facts? Searching for Fact-checked
  Information to Alleviate the Spread of Fake News},'' \emph{Proceedings of the
  Conference on Empirical Methods in Natural Language Processing}, pp.
  7717--7731, November 2020, {Virtual}.

\bibitem{fung2021infosurgeon}
Y.~Fung, C.~Thomas, R.~Gangi~Reddy, S.~Polisetty, H.~Ji, S.-F. Chang,
  K.~McKeown, M.~Bansal, and A.~Sil, ``{{I}nfo{S}urgeon: Cross-Media
  Fine-grained Information Consistency Checking for Fake News Detection},''
  \emph{Proceedings of the Annual Meeting of the Association for Computational
  Linguistics and the International Joint Conference on Natural Language
  Processing}, vol.~1, pp. 1683--1698, August 2021, {Virtual}.

\bibitem{lee2018stacked}
K.-H. Lee, X.~Chen, G.~Hua, H.~Hu, and X.~He, ``{Stacked Cross Attention for
  Image-Text Matching},'' \emph{Proceedings of the European Conference on
  Computer Vision}, p. 212–228, September 2018, {Munich, Germany}.

\bibitem{xu2020crossmodal}
X.~Xu, T.~Wang, Y.~Yang, L.~Zuo, F.~Shen, and H.~T. Shen, ``{Cross-Modal
  Attention With Semantic Consistence for Image-Text Matching},'' \emph{IEEE
  Transactions on Neural Networks and Learning Systems}, vol.~31, no.~12, pp.
  5412--5425, February 2020.

\bibitem{zhang2018deep}
Y.~Zhang and H.~Lu, ``{Deep Cross-Modal Projection Learning for Image-Text
  Matching},'' \emph{Proceedings of the European Conference on Computer
  Vision}, p. 707–723, September 2018, {Munich, Germany}.

\bibitem{li2019visual}
K.~Li, Y.~Zhang, K.~Li, Y.~Li, and Y.~Fu, ``{Visual Semantic Reasoning for
  Image-Text Matching},'' \emph{Proceedings of the IEEE/CVF International
  Conference on Computer Vision}, pp. 4653--4661, October 2019, {Seoul, Korea
  (South)}.

\bibitem{li2020oscar}
X.~Li, X.~Yin, C.~Li, P.~Zhang, X.~Hu, L.~Zhang, L.~Wang, H.~Hu, L.~Dong,
  F.~Wei, Y.~Choi, and J.~Gao, ``{Oscar: Object-Semantics Aligned Pre-training
  for Vision-Language Tasks},'' \emph{Proceedings of the European Conference on
  Computer Vision}, pp. 121--137, August 2020, {Glasgow, United Kingdom}.

\bibitem{tan2020detecting}
R.~Tan, B.~A. Plummer, and K.~Saenko, ``{Detecting Cross-Modal Inconsistency to
  Defend Against Neural Fake News},'' \emph{arXiv preprint arXiv:2009.07698},
  October 2020.

\bibitem{mccrae2021multi}
S.~McCrae, K.~Wang, and A.~Zakhor, ``{Multi-Modal Semantic Inconsistency
  Detection in Social Media News Posts},'' \emph{arXiv preprint
  arXiv:2105.12855}, May 2021.

\bibitem{iuliani2019video}
M.~Iuliani, D.~Shullani, M.~Fontani, S.~Meucci, and A.~Piva, ``{A Video
  Forensic Framework for the Unsupervised Analysis of MP4-Like File
  Container},'' \emph{IEEE Transactions on Information Forensics and Security},
  vol.~14, no.~3, pp. 635--645, March 2019.

\bibitem{raissi2002theory}
R.~Raissi, ``{The Theory Behind MP3},''
  \url{http://citeseerx.ist.psu.edu/viewdoc/summary?doi=10.1.1.113.6804}, 2002.

\bibitem{guera2019we}
D.~G\"{u}era, S.~Baireddy, P.~Bestagini, S.~Tubaro, and E.~Delp, ``{We Need No
  Pixels: Video Manipulation Detection Using Stream Descriptors},''
  \emph{Proceedings of the International Conference on Machine Learning,
  Synthetic-Realities: Deep Learning for Detecting AudioVisual Fakes Workshop},
  June 2019, {Long Beach, CA}.

\bibitem{vazquez2020video}
D.~V\'{a}zquez-Pad\'{i}n, M.~Fontani, D.~Shullani, F.~P\'{e}rez-Gonz\'{a}lez,
  A.~Piva, and M.~Barni, ``{Video Integrity Verification and GOP Size
  Estimation Via Generalized Variation of Prediction Footprint},'' \emph{IEEE
  Transactions on Information Forensics and Security}, vol.~15, pp. 1815--1830,
  November 2020.

\bibitem{iso-mp4file}
``{ISO/IEC 14496-14:2003 - Information Technology -- Coding of Audio-Visual
  Objects -- Part 14: MP4 File Format},''
  \url{https://www.iso.org/standard/38538.html}.

\bibitem{yang2020efficient}
P.~Yang, D.~Baracchi, M.~Iuliani, D.~Shullani, R.~Ni, Y.~Zhao, and A.~Piva,
  ``{Efficient Video Integrity Analysis Through Container Characterization},''
  \emph{IEEE Journal of Selected Topics in Signal Processing}, vol.~14, no.~5,
  pp. 947--954, July 2020.

\bibitem{xiang2021forensic}
Z.~Xiang, J.~Horv\'{a}th, S.~Baireddy, P.~Bestagini, S.~Tubaro, and E.~J. Delp,
  ``{Forensic Analysis of Video Files Using Metadata},'' \emph{Proceedings of
  the IEEE/CVF Conference on Computer Vision and Pattern Recognition
  Workshops}, pp. 1042--1051, June 2021, {Virtual}.

\bibitem{su2009asource}
Y.~Su, J.~Xu, and B.~Dong, ``{A Source Video Identification Algorithm Based on
  Motion Vectors},'' \emph{Proceedings of the Second International Workshop on
  Computer Science and Engineering}, vol.~2, pp. 312--316, October 2009,
  {Qingdao, China}.

\bibitem{yao2020double}
H.~Yao, R.~Ni, and Y.~Zhao, ``{Double Compression Detection for H. 264 videos
  with Adaptive GOP Structure},'' \emph{Multimedia Tools and Applications},
  vol.~79, no.~9, pp. 5789--5806, 2020.

\bibitem{fang2019detection}
Q.~Fang, X.~Jiang, T.~Sun, Q.~Xu, and K.~Xu, ``{Detection of HEVC Double
  Compression with Different Quantization Parameters Based on Property of DCT
  Coefficients and TUs},'' \emph{Proceedings of the International Congress on
  Image and Signal Processing, BioMedical Engineering and Informatics}, pp.
  1--6, October 2019.

\bibitem{he2021framewise}
P.~He, H.~Li, H.~Wang, S.~Wang, X.~Jiang, and R.~Zhang, ``{Frame-Wise Detection
  of Double HEVC Compression by Learning Deep Spatio-Temporal Representations
  in Compression Domain},'' \emph{IEEE Transactions on Multimedia}, vol.~23,
  pp. 3179--3192, September 2020.

\bibitem{jiang2020detection}
X.~Jiang, Q.~Xu, T.~Sun, B.~Li, and P.~He, ``{Detection of HEVC Double
  Compression With the Same Coding Parameters Based on Analysis of Intra Coding
  Quality Degradation Process},'' \emph{IEEE Transactions on Information
  Forensics and Security}, vol.~15, pp. 250--263, May 2019.

\bibitem{altinisik2022camera}
E.~Altinisik and H.~T. Sencar, ``{Camera Model Identification Using Container
  and Encoding Characteristics of Video Files},'' \emph{arXiv preprint
  arXiv:2112.07945}, February 2022.

\bibitem{qiao2013improved}
M.~Qiao, A.~H. Sung, and Q.~Liu, ``{Improved Detection of {MP3} Double
  Compression Using Content-Independent Features},'' \emph{Proceedings of the
  IEEE International Conference on Signal Processing, Communication and
  Computing}, pp. 1--4, August 2013, {KunMing, China}.

\bibitem{ma2014detecting}
P.~Ma, R.~Wang, D.~Yan, and C.~Jin, ``{Detecting Double-Compressed {MP3} with
  the Same Bit-Rate},'' \emph{Journal of Software}, vol.~9, no.~10, pp.
  2522--2527, October 2014.

\bibitem{ma2014huffman}
P.~Ma, R.~Wang, D.~Yan, and C.~Jin, ``{A Huffman Table Index Based Approach to
  Detect Double {MP3} Compression},'' in \emph{Digital-Forensics and
  Watermarking}.\hskip 1em plus 0.5em minus 0.4em\relax Springer, July 2014,
  pp. 258--271.

\bibitem{yan2018compression}
D.~Yan, R.~Wang, J.~Zhou, C.~Jin, and Z.~Wang, ``{Compression History Detection
  for {MP3} Audio},'' \emph{KSII Transactions on Internet and Information
  Systems}, vol.~12, no.~2, pp. 662--675, February 2018.

\bibitem{bianchi2014detection}
T.~Bianchi, A.~De~Rosa, M.~Fontani, G.~Rocciolo, and A.~Piva, ``{Detection and
  Localization of Double Compression in MP3 Audio Tracks},'' \emph{EURASIP
  Journal on information Security}, vol. 2014, no.~10, May 2014.

\bibitem{jin2016efficient}
C.~Jin, R.~Wang, D.~Yan, P.~Ma, and J.~Zhou, ``{An Efficient Algorithm for
  Double Compressed AAC Audio Detection},'' \emph{Multimedia Tools and
  Applications}, vol.~75, no.~8, pp. 4815--4832, April 2016.

\bibitem{huang2018aac}
Q.~Huang, R.~Wang, D.~Yan, and J.~Zhang, ``{AAC Audio Compression Detection
  Based on QMDCT Coefficient},'' in \emph{{Cloud Computing and
  Security}}.\hskip 1em plus 0.5em minus 0.4em\relax Springer International
  Publishing, September 2018, pp. 347--359.

\bibitem{huang2018aac2}
Q.~Huang, R.~Wang, D.~Yan, and J.~Zhang, ``{AAC Double Compression Audio
  Detection Algorithm Based on the Difference of Scale Factor},''
  \emph{Information}, vol.~9, no.~7, p. 161, July 2018.

\bibitem{hosler2021do}
B.~Hosler, D.~Salvi, A.~Murray, F.~Antonacci, P.~Bestagini, S.~Tubaro, and
  M.~C. Stamm, ``{Do Deepfakes Feel Emotions? A Semantic Approach to Detecting
  Deepfakes Via Emotional Inconsistencies},'' \emph{Proceedings of the IEEE/CVF
  Conference on Computer Vision and Pattern Recognition Workshops}, pp.
  1013--1022, June 2021, {Virtual}.

\end{thebibliography}
}
\end{document}